\pdfoutput=1

\documentclass[11pt]{article}

\usepackage[final]{acl}

\usepackage{newunicodechar}
\newunicodechar{ }{\hspace{1em}}
\usepackage{times}
\usepackage{latexsym}
\usepackage{epigraph}  
\usepackage{amsmath}
\usepackage{enumitem}
\usepackage{ulem}
\usepackage{multirow}
\usepackage{booktabs}

\usepackage[T1]{fontenc}

\usepackage[utf8]{inputenc}

\usepackage{microtype}

\usepackage{inconsolata}

\usepackage{graphicx}

\usepackage[ruled,vlined,linesnumbered]{algorithm2e}

\title{MemTool: Optimizing Short-Term Memory Management for\\ Dynamic Tool Calling in LLM Agent Multi-Turn Conversations}

\author{
 \textbf{Elias Lumer}, 
 \textbf{Anmol Gulati}, 
 \textbf{Vamse Kumar Subbiah}, \\
 \textbf{Pradeep Honaganahalli Basavaraju} 
 \textbf{and James A. Burke} \\
 \textit{Commercial Technology and Innovation Office, PricewaterhouseCoopers, U.S.A}
}

\begin{document}
\maketitle
\begin{abstract}
Large Language Model (LLM) agents have shown significant autonomous capabilities in dynamically searching and incorporating relevant tools or Model Context Protocol (MCP) servers for individual queries. However, fixed context windows limit effectiveness in multi-turn interactions requiring repeated, independent tool usage. We introduce MemTool, a short-term memory framework enabling LLM agents to dynamically manage tools or MCP server contexts across multi-turn conversations. MemTool offers three agentic architectures: 1) Autonomous Agent Mode, granting full tool management autonomy, 2) Workflow Mode, providing deterministic control without autonomy, and 3) Hybrid Mode, combining autonomous and deterministic control. Evaluating each MemTool mode across 13+ LLMs on the ScaleMCP benchmark, we conducted experiments over 100 consecutive user interactions, measuring tool removal ratios (short-term memory efficiency) and task completion accuracy. In Autonomous Agent Mode, reasoning LLMs achieve high tool-removal efficiency (90–94\% over a 3-window average), while medium-sized models exhibit significantly lower efficiency (0–60\%). Workflow and Hybrid modes consistently manage tool removal effectively, whereas Autonomous and Hybrid modes excel at task completion. We present trade-offs and recommendations for each MemTool mode based on task accuracy, agency, and model capabilities.
\end{abstract}

\section{Introduction}\label{sec:Introduction}
\epigraph{
  \textit{The LLM is the CPU, and its context window is the RAM, representing the LLM's working memory.}\footnotemark
}{}                 

\footnotetext{%
  Adapted from a talk by Andrej Karpathy, “Software in the Era of AI,” presented by
  Y Combinator, AI Startup School, 17 June 2025.\cite{karpathy2025keynote}.
}

Recent breakthroughs in Large Language Model (LLM) agents have enabled powerful and scalable agentic systems which autonomously search, discover, equip, and use a dynamic repository of tools or MCP servers \cite{modelcontext2025tools}. Dynamic tool calling, also known as function calling, allows LLM agents to interact with external APIs or systems, while not being constrained to a fixed set of tools when the agent is initialized \cite{lumer2025scalemcpdynamicautosynchronizingmodel}. This breakthrough enables LLM agents to go beyond a fixed number of tools, granting them the autonomy to explore, identify, and integrate new tools based off the user query, much like a human navigating a mobile app store. Moreover, as LLM agents become increasingly embedded in user session-based (chat, voice, video) applications, managing the limited context window of the model is necessary for multi-turn conversations. Known as context engineering and memory management, these systems allow multi-turn user interactions to persist not only across a single session (short-term memory) but also across multiple long-term sessions (long-term memory) \cite{schmid2025context,langchain2025context}.

Despite the advancements in dynamic tool-using agents and user-agent chat-based memory, there is still a significant gap in how LLM agents manage its short-term memory of tools in multi-turn dynamic tool retrieval settings. LLM agents can discover and add hundreds of new tools or MCP servers its context window, but need to reduce the number of available tools in its context upon solving the user question and no longer needing them. One primary concern is that the current literature on short-term memory management predominantly focuses on compressing the context between an assistant and a user through summarization and truncation of messages \cite{wu2025humanmemoryaimemory,breunig2025context}. However, these methods apply primarily to conversational tasks, where LLM agents perform a sequence of actions requested by the user, and do not address the dynamic retrieval and management of tools within an agent's context window.

\begin{figure*}[t]
  \centering
  \includegraphics[width=\textwidth]{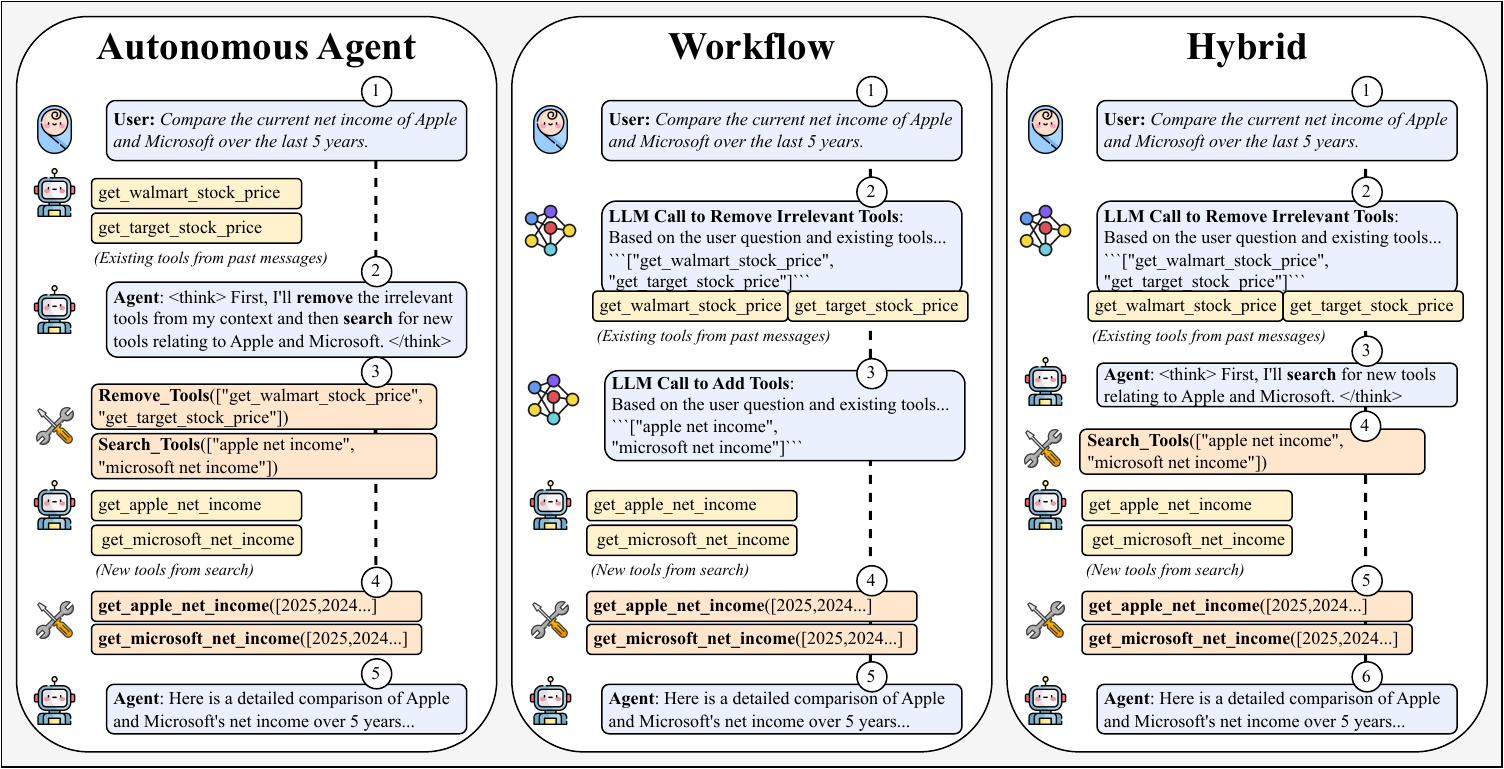} 
  \caption{MemTool three modes architecture and end-to-end walkthrough of user-assistant interaction across multi-turn chats. In the example, the previous user session continues after the user asked about Walmart and Target's stock price, given the existing tools still in the context window of the LLM agent. The LLM agent (robot icon) represents an LLM binded with a set of tools, where the LLM call (neural net icon) represents a fresh LLM chat completion message with no tools.}
  \label{fig:memtool_modes}
\end{figure*}
 
A second concern is the limited research on multi-turn scenarios among LLM agents with dynamic tool retrieval. Previous work has proven that LLM agents or agentic workflows can handle thousands of tools or MCPs without a significant drop in accuracy, due to only passing in relevant tools to the LLM agent by retrieval augmented generation \cite{icaart25,chen2024reinvoketoolinvocationrewriting}. However, the existing state-of-the-art methods do not address the removal and management of tools (along with the addition of tools) within the LLM agent context window across a 100+ multi-turn session, while using native function calling \cite{fei2025mcpzeroactivetooldiscovery,lumer2025scalemcpdynamicautosynchronizingmodel}. To effectively use dynamic tool retrieval agents in production environments, there needs to be thorough evaluation of how tools persist through multi-turn conversations in the same session. While some prior work has touched on long-term memory and personalization of tool-use \cite{zhang2025personaagentlargelanguagemodel,hao2025evaluatingpersonalizedtoolaugmentedllms,cheng2025toolspectrumpersonalizedtool}, there remains a gap in how LLM agents manage their short-term tool memory through a multi-turn session.

In this paper, we introduce MemTool, a short-term memory framework for multi-turn dynamic tool use agents, enabling agents to manage their own context window of hundreds and thousands of searchable tools or MCP servers. MemTool is comprised of three modes--containing different levels of agency or autonomy for the LLM--for managing an LLM agent's context window of tools across multi-turn conversations (See Figure~\ref{fig:memtool_modes}). The first mode is the Autonomous Agent Mode, which gives full control to the LLM agent to add and remove tools from its context window (through function calling) while simultanously answering the user question. The second mode is a predefined agentic workflow; first always removes irrelevant tools from its context, then selects additional tools through semantic or lexical search in a tool knowledge base, finally giving the LLM agent only the new tools to answer the user's question. The third mode is a hybrid of autonomous agent and workflow, which first starts by always removing tools from its context, then giving the LLM agent autonomy to only add dynamic tools and use them to answer the user question. 

After evaluating the three MemTool modes on the ScaleMCP benchmark of 5,000 MCP servers, we find that Autonomous Agent Mode achieves the highest tool-removal efficiency (90–94\%) with reasoning LLMs, while Workflow and Hybrid Modes offer consistently effective tool removal across all models. In contrast, Autonomous Agent and Hybrid Modes demonstrate the highest task completion rates, leveraging agentic corrective loops for dynamic tool retrieval. Finally, we present trade-offs and recommendations for each MemTool mode based on task accuracy, agency, and model capabilities.

\section{Background}\label{sec:background}

\subsection{Memory for LLM agents}\label{subsec:memory}
Memory for Large Language Model (LLM) agents enables the persistent accumulation of knowledge, sustained contextual awareness, and adaptive decision-making informed by historical interactions and prior experiences. Existing literature divides memory for AI systems into short-term and long-term memory \cite{wu2025humanmemoryaimemory}. For chat-based applications, short-term memory is often described for memories that exist within the session, and long-term memory enables memories beyond the session.
\subsubsection{Short-Term Memory}
The majority of flagship model context windows fall between 100,000 to 1,000,000 tokens \cite{vellum2025llm,shang2025longrope2nearlosslessllmcontext}. Short-term memory targets all input tokens within the limited LLM context window. It is often broken down into sensory memory and working memory-where sensory memory encapsulates all multi-modal inputs and tool calls, and working memory consists of the system, assistant, and human messages or chain of thoughts that occur within the session \cite{wu2025humanmemoryaimemory}. Previous work on short-term memory is aimed at multi-turn conversations, which prevent the overflow of tokens within a model's context window, as well as prune necessary context that is found to distract the model during a task \cite{laban2025llmslostmultiturnconversation,hong2025context}. The context engineering strategies for short-term memory across multi-turn dialogues involve summarization or truncation of previous chat messages \cite{chirkova2025provenceefficientrobustcontext,breunig2025context,schmid2025context,langchain2025context,alake2025agentmemory}. However, it is noted that summarizing long multi-turn dialogues is prone to over-summarization or information loss \cite{ravaut2024contextutilizationsummarizationlarge,
wang2025recursivelysummarizingenableslongterm,
maharana2024evaluatinglongtermconversationalmemory,
laban2025llmslostmultiturnconversation,
pan2025memoryconstructionretrievalpersonalized,
wu2025longmemevalbenchmarkingchatassistants,
shan2025cognitivememorylargelanguage}. In our paper, MemTool fits within short-term memory for LLM agents, that manage and optimize the available tool context of the model across multi-turn chats within a session. MemTool is the first study to address and evaluate short-term memory for dynamic tool-use across multi-turn conversations, while providing three architectures (Figure~\ref{fig:memtool_modes}) to optimize and manage available tools within the context window.

\subsubsection{Long-Term Memory}
Long-term memory for LLMs involves storing information for extended periods that persist across sessions, enabling the model to retain knowledge and learned abilities over time \cite{wu2025humanmemoryaimemory,park2023generativeagentsinteractivesimulacra,zhong2023memorybankenhancinglargelanguage}. Two types of long-term memory are explicit memory and implicit memory. Explicit memory consists of the recollection of key facts and events; implicit memory consists of learned skills, procedures, and long tasks. Explicit memories are further divided into episodic memory and semantic memory. Episodic memory refers to personal experiences and events, where semantic memory refers to facts and knowledge about the environment. Episodic memories are commonly used in chat applications that involve user personalization over time (e.g. knowing the user prefers a brand of clothing) \cite{alake2025agentmemory}. 

Existing long‑term memory frameworks include \cite{mem02025}, \cite{zep2025}, and \cite{letta2025,packer2024memgptllmsoperatingsystems}, which manage the storage and personalization of LLM agent long‑term interactions with a user. In addition, large model providers that serve their model to users also store long‑term memories across sessions for their users \cite{openai2025,google2025,perplexity2025}.

Recent works have aimed at long-term memory for tool usage, that adapt the tool-use preferences of the user over time \cite{zhang2025personaagentlargelanguagemodel,hao2025evaluatingpersonalizedtoolaugmentedllms,cheng2025toolspectrumpersonalizedtool}. MemTool is complementary to the aforementioned work, as long-term memory and short-term memory both serve as critical elements of context engineering for tools. Furthermore, some methods of updating or adding memories for tool personalization include an LLM agent calling a tool to update memories \cite{letta2025,xu2025amemagenticmemoryllm}. In our work, we demonstrate that certain LLM models perform better than others in managing their short-term memory context while simultaneously solving the user query (Table~\ref{tab:main_results}). MemTool's three modes (Figure~\ref{fig:memtool_modes}) can be partially extrapolated to how LLM agents manage their long-term memory as well.

\subsection{Tool Selection and Retrieval for LLM Agents}\label{subsec:dynamic_tool_llms}

Large language models inherently face limitations regarding the number of tools or functions they can concurrently invoke or manage. Complex, multi-step tool interactions place substantial demands on the LLM's reasoning processes, complicating tool selection and sequencing. Additionally, leading model providers  impose varying constraints on the number of tools a single LLM API request can have, within the range of 128-512 \cite{openaifunctioncalling2024}. To circumvent these limitations, recent studies \cite{lumericaart25,chen2024reinvoketoolinvocationrewriting} have introduced advanced retrieval-augmented generation (RAG) strategies. These approaches do not rely on model fine-tuning but instead store large toolsets externally within vector databases or structured knowledge graphs \cite{lumer2025graphragtoolfusion,peng2024graphretrievalaugmentedgenerationsurvey}, dynamically selecting and equipping only the necessary tools during runtime. In MemTool, we also rely on out-of-the-box embeddings for the 5,000 MCP servers stored in our tool knowledge base. However, to increase retrieval accuracy further, previous work have fine-tuned embedders for the tool knowledge base \cite{wu2024sealtoolsselfinstructtoollearning,qin2023toolllmfacilitatinglargelanguage,anantha2023protipprogressivetoolretrieval,yuan2024craftcustomizingllmscreating,zheng2024toolrerankadaptivehierarchyawarereranking}.

An alternative methodology utilizes agentic RAG frameworks, which give LLM agents with specialized tool-discovery tools that autonomously identify and invoke appropriate tools as needed \cite{singh2025agenticretrievalaugmentedgenerationsurvey,li2023apibankcomprehensivebenchmarktoolaugmented,du2024anytoolselfreflectivehierarchicalagents}. This represents a departure from static, predefined retrieval setups \cite{lumer2024toolshedscaletoolequippedagents,chen2024reinvoketoolinvocationrewriting}. However, previous research highlighted challenges, particularly with earlier GPT architectures that struggled to leverage these dynamic search functionalities effectively \cite{li2023apibankcomprehensivebenchmarktoolaugmented}. Recent contributions and stronger tool calling LLM models \cite{lumer2025scalemcpdynamicautosynchronizingmodel,fei2025mcpzeroactivetooldiscovery} significantly advance this domain by enabling an agent to dynamically query a comprehensive repository of over 5,000 MCP servers, facilitating efficient tool incorporation into the agent’s context window. However, these methods do not address the removal and management of dynamic tools for a 100+ multi-turn session. In our paper, we address this gap in short-term tool memory research, propose three novel modes or architectures to enable LLM agents to manage their tool context, and evaluate the three modes on over 10+ state-of-the-art LLM models. For our experiments, we use the ScaleMCP benchmark of 5,000 MCP servers \cite{lumer2025scalemcpdynamicautosynchronizingmodel} with out-of-the-box to store and retrieve tool names and descriptions.

\begin{algorithm}[t]
\footnotesize
\caption{Autonomous Agent Mode}\label{alg:agent}
\KwIn{LLM, user query $q$, prior messages $\text{msgs}_{\mathrm{prev}}$, 
      prior tool set $T_{\mathrm{prev}}$, vectorDB, 
      top-$k{=}5$, tool-limit $L{=}128$, 
      history manager $H\!\in\!\{\textsc{TruncateHistory},\textsc{SummarizeHistory}\}$}
\KwOut{final answer $a$}

\textbf{messages} $\leftarrow H(\text{msgs}_{\mathrm{prev}}) + q$\; \par
\texttt{T} $\leftarrow T_{\mathrm{prev}} \cup \{$\par
\hspace*{2em}\texttt{Search\_Tools(keywords: List[str])},\par
\hspace*{2em}\texttt{Remove\_Tools(tool\_names: List[str])}\par
$\}$
\While{true}{
  $\mathsf{reply} \leftarrow LLM(\text{messages},\text{tools}=T)$\;
  \eIf{$\mathsf{reply}.\text{type} = \texttt{tool\_call}$}{
    \Switch{$\mathsf{reply}.\text{tool}$}{
      \Case{\texttt{Remove\_Tools}}{
        $R \leftarrow \mathsf{reply}.\text{args}$\;
        \texttt{T} $\leftarrow T \setminus (R \setminus$%
\par
\hspace*{2em}$\{\texttt{Search\_Tools},\ \texttt{Remove\_Tools}\})$\;
      }
      \Case{Search\_Tools}{
        $Q \leftarrow \mathsf{reply}.\text{args}$\;
        $S \leftarrow \texttt{VectorSearch}(Q,\mathcal{V},k{=}5)$\;
        $T \leftarrow T \cup S$\;
        \If{$|T| > L$}{\textbf{Raise} \texttt{Error("Hit $L$ limit, remove tools.")}\;}
      }
      \Other{
        $\text{result} \leftarrow \texttt{ExecuteTool}(\mathsf{reply})$\;
        append $(\mathsf{reply}.\text{tool},\text{result})$ to $\text{messages}$\;
      }
    }
  }{
    append $\mathsf{reply}$ to $\text{messages}$\;
    $a \leftarrow \mathsf{reply}.\text{content}$\;
    \Return $a$\;
  }
}
\end{algorithm}

\subsection{Tool Calling or LLM Invocation}\label{subsec:llm_invocation_of_tools}

Beyond selecting and retrieving relevant tools, another critical aspect is the actual invocation of these tools by LLM agents. Several studies focus specifically on enhancing LLM capabilities for precise and effective tool invocation \citep{hao2024toolkengptaugmentingfrozenlanguage,qin2023toolllmfacilitatinglargelanguage,patil2023gorillalargelanguagemodel}. Recent advancements in fine-tuning methods have contributed significantly to improving LLMs' proficiency in tool calling, including MOLoRA's modular approach \citep{hao2024citienhancingtoolutilizing}, efficient tree-structured techniques \citep{zhu2025dividethenaggregateefficienttoollearning}, and carefully constructed datasets generated collaboratively by multi-agent frameworks \citep{liu2024toolacewinningpointsllm,zhuang2025hephaestusimprovingfundamentalagent}.

Despite the demonstrated benefits of fine-tuning strategies, our work deliberately adopts a different path, emphasizing plug-and-play methodologies compatible with out-of-the-box models and standard embedding solutions from providers such as OpenAI, Google, Anthropic, and Meta \citep{openai_2025,google_gemini_2025,anthropic_2025,meta_llama_2025}. Specifically, we investigate whether standard, commercially available LLMs can autonomously manage their tool context dynamically—removing tools when they lose relevance and incorporating new tools on demand. To this end, our framework equips the LLM with explicit operations for adding and removing tools from its active context, thus enabling greater autonomy in tool management. Additionally, acknowledging diverse operational requirements, we introduce two alternative execution modes designed to reduce autonomy in favor of more deterministic and controlled tool invocation behaviors.

\SetKwSwitch{Switch}{Case}{Other}{}{case}{otherwise}{}{}
\DontPrintSemicolon

\section{Method}
\subsection{MemTool}
MemTool enables an LLM agent to manage its own context window of dynamic tools across multi-turn sessions. The broader implication of MemTool is that LLM agents can operate in production environments with a non-fixed set of tools at its disposal, searching, equipping, and removing tools or MCP servers, similarly to a human navigating a mobile app store. Specifically, we propose three modes or architectures (Figure~\ref{fig:memtool_modes}) that grant varying degrees of autonomy to the LLM agent to optimize its short-term memory context window of tools (autonomous agent, workflow, and hybrid). We note trade-offs and recommendations for each mode, in each subsequent section.

\subsubsection{Autonomous Agent Mode}
MemTool Autonomous Agent Mode grants full autonomy to the LLM agent to manage its context window of available tools while simultaneuously answering the user task, across multi-turn conversations. The manner in which tools or MCP servers are removed or searched for is handled by two additional tools equipped to the LLM agent. The system prompt for the autonomous agent explains its task is to answer the user question by searching for available tools while also removing irrelevant tools it no longer needs. The full system prompt is in Appendix~\ref{fig:agent_system_prompt}. Firstly, the agent is given a search tool to add tools to its context window. This tool is adapted from ScaleMCP \cite{lumer2025scalemcpdynamicautosynchronizingmodel}. Secondly, the agent is given a remove tool, which is the critical element for managing over 100+ user questions sequentially. If the LLM agent calls the remove tool with a list of tools it has access to that are no longer relevant for the user question, we automatically remove them from the API call (for example, "tools=tools" in chat completions API) \cite{openaifunctioncalling2024}. We append any remove and/or search tool calls and tool call results to the context window, with helpful result messages such as "N tools added: {tool names}" or "N tools removed: {tool names}". Finally, the LLM agent will use its new tools to answer the user question, which the messages and existing are persisted to additional queries within the session. 

In Algorithm~\ref{alg:agent}, we first prune the previous messages either by truncation or summarization, in case the LLM agent used too many tokens in the previous query. This truncation does not affect the available tools, only dialogue messages. Then, the LLM agent enters a while loop, calling tools described above to manage its short-term memory context window of tools while answering the user question. Notably, when the SearchTool adds tools to the LLM agent's context, and the total exceeds the LLM API's tool limit, we return an error message and prompt the agent to remove tools from its context.

\begin{algorithm}[t]
\footnotesize
\caption{Workflow Mode}
\label{alg:workflow}
\KwIn{LLM, user query $q$, prior messages $\text{msgs}_{\mathrm{prev}}$, 
      prior tool set $T_{\mathrm{prev}}$, vectorDB $\mathcal{V}$, 
      top-$k{=}5$, tool-limit $L{=}128$, 
      history manager $H\!\in\!\{\textsc{TruncateHistory},\textsc{SummarizeHistory}\}$}
\KwOut{final answer $a$}

\textbf{messages} $\leftarrow H(\text{msgs}_{\mathrm{prev}}) + q$\;
\texttt{T} $\leftarrow T_{\mathrm{prev}}$\;

$\mathsf{R} \leftarrow LLM_{\mathrm{prune}}(\text{messages},\texttt{T})$\;
\texttt{T} $\leftarrow \texttt{T} \setminus \mathsf{R}$\;

$\mathsf{Q} \leftarrow LLM_{\mathrm{search}}(\text{messages},q)$\;
\If{$\mathsf{Q} \neq \emptyset$}{
  $S \leftarrow \texttt{VectorSearch}(\mathsf{Q},\mathcal{V},k{=}5)$\;
  \texttt{T} $\leftarrow \texttt{T} \cup S$\;
  
  \While{$|\texttt{T}| > L$}{
    $\mathsf{R} \leftarrow LLM_{\mathrm{prune}}(\text{messages},\texttt{T})$\;
    \If{$\mathsf{R} = \emptyset$}{
      \textbf{Raise} \texttt{Error("$T$>$L$")}\;
    }
    \texttt{T} $\leftarrow \texttt{T} \setminus \mathsf{R}$\;
  }
} 

\While{true}{
  $\mathsf{reply} \leftarrow LLM(\text{messages},\text{tools}=\texttt{T})$\;
  \eIf{$\mathsf{reply}.\text{type} = \texttt{tool\_call}$}{
    $\text{result} \leftarrow \texttt{ExecuteTool}(\mathsf{reply})$\;
    append $(\mathsf{reply}.\text{tool},\text{result})$ to \textbf{messages}\;
  }{
    append $\mathsf{reply}$ to \textbf{messages}\;
    $a \leftarrow \mathsf{reply}.\text{content}$\;
    \Return $a$\;
  }
}
\end{algorithm}

\paragraph{Recommendations for Autonomous Agent Mode.} If opting to use Autonomous Agent Mode for MemTool's short-term tool memory:
\begin{itemize}[nosep]
  \item Certain LLM models fail to remove irrelevant tools from its context, causing the number of tools to increase until an upper limit.
  \item Prompting the system message of the LLM agent can provide detailed guidelines on when to remove tools and when to add them.
  \item Upper limit tool errors (e.g. 128) 'encourage' LLMs to remove tools. 
  \item Including the current tool count (as a dynamic variable) in the system message helped the agent remove tools, rather than letting it infer from the tools equipped to the LLM from the API.
\end{itemize}
\subsubsection{Workflow Mode}
MemTool Workflow Mode limits autonomy to the LLM agent by abstracting the dynamic tool management to a fixed workflow that occurs after every user query. Instead, two fresh LLM calls occur in sequence to remove irrelevant tools (to the user query) and search for new tools to add. The irrelevant tools were once relevant in previous chat messages within the session, but no longer relevant. This Workflow Mode is commonly seen in state-of-the-art tool retrieval approaches \cite{lumericaart25,chen2024reinvoketoolinvocationrewriting}. The novel differentiator that MemTool Workflow adds to previous work is enabling multi-turn conversations, adding a tool pruning LLM call, and by persisting the tools across the session. 

\begin{algorithm}[t]
\footnotesize
\caption{Hybrid Mode}
\label{alg:hybrid}
\KwIn{LLM, user query $q$, prior messages $\text{msgs}_{\mathrm{prev}}$, 
      prior tool set $T_{\mathrm{prev}}$, vectorDB $\mathcal{V}$, 
      top-$k{=}5$, tool-limit $L{=}128$, 
      history manager $H\!\in\!\{\textsc{TruncateHistory},\textsc{SummarizeHistory}\}$}
\KwOut{final answer $a$}

\textbf{messages} $\leftarrow H(\text{msgs}_{\mathrm{prev}}) + q$\;
\texttt{T} $\leftarrow T_{\mathrm{prev}}$\;

$\mathsf{R} \leftarrow LLM_{\mathrm{prune}}(\text{messages},\texttt{T})$\;
\texttt{T} $\leftarrow \texttt{T} \setminus \mathsf{R}$\;

\texttt{T} $\leftarrow T_{\mathrm{prev}} \cup \{$\par
\hspace*{2em}\texttt{Search\_Tools(keywords: List[str])}$\}$

\While{true}{
  $\mathsf{reply} \leftarrow LLM(\text{messages},\text{tools}=\texttt{T})$\;
  \eIf{$\mathsf{reply}.\text{type} = \texttt{tool\_call}$}{
    \Switch{$\mathsf{reply}.\text{tool}$}{
      \Case{\texttt{Search\_Tools}}{
        $Q \leftarrow \mathsf{reply}.\text{args}$\;
        $S \leftarrow \texttt{VectorSearch}(Q,\mathcal{V},k{=}5)$\;
        \texttt{T} $\leftarrow \texttt{T} \cup S$\;
        \While{$|\texttt{T}| > L$}{
          $\mathsf{R} \leftarrow LLM_{\mathrm{prune}}(\text{messages},\texttt{T})$\;
          \If{$\mathsf{R} = \emptyset$}{
            \textbf{Raise} \texttt{Error("$T$>$L$.")}\;
          }
          \texttt{T} $\leftarrow \texttt{T} \setminus \mathsf{R}$\;
        }
      }
      \Other{
        $\text{result} \leftarrow \texttt{ExecuteTool}(\mathsf{reply})$\;
        append $(\mathsf{reply}.\text{tool},\text{result})$ to \textbf{messages}\;
      }
    }
  }{
    append $\mathsf{reply}$ to \textbf{messages}\;
    $a \leftarrow \mathsf{reply}.\text{content}$\;
    \Return $a$\;
  }
}
\end{algorithm}

As seen in Algorithm~\ref{alg:workflow}, the same messages and previous tools are initialized as the agent mode. Then, a fresh LLM call examines the current tool list along with the user query and decides to remove tools no longer relevant. Then, a fresh LLM examines the available tools and user query and decides to search for new tools if the existing ones are not sufficient. After removing and adding tools to the existing tool list, the LLM agent is finally initialized in the while loop, identical to the agent mode, less the ability to call SearchTool and RemoveTool.
\paragraph{Recommendations for Workflow Mode.} If opting to use Workflow Mode for MemTool's short-term tool memory:
\begin{itemize}[nosep]
  \item The simplicity of the workflow is a recommended first choice, however, once the LLM agent is initialized with the tool list, there is no mechanism to revert back to searching new tools.
  \item Limiting agency is seen to prevent any self-correctness. While previous works \cite{lumericaart25} have suggested to add a self-correctness step to the workflow, these patterns get overly complex and restrictive.
\end{itemize}
\subsubsection{Hybrid Mode}
MemTool Hybrid Mode grants the LLM agent partial autonomy while restricting others. After observing that LLMs tend to struggle with tool removal (the pruning aspect of short-term tool memory management) but perform well in searching and adding tools, Hybrid Mode combines both approaches to leverage their respective strengths. Hybrid Mode includes the fresh LLM call to remove no longer relevant tools from the context window, while granting the LLM agent autonomy to use the SearchTool to add and use a dynamic set of tools. This partial agency solves the limitation found in \cite{lumericaart25} of complex self-correcting workflows. 

\begin{table*}[t]
\centering
\footnotesize
\begin{tabular}{llrrrr}
\toprule
\textbf{Mode} & \textbf{LLM} & \textbf{Avg Removal Ratio 3T} & \textbf{Avg Residual 3T} & \textbf{Tool Correctness} & \textbf{Task Completion} \\
\midrule
\multirow{13}{*}{\shortstack{\textbf{Autonomous} \\ \textbf{Agent}}}
 & GPT-o3            & \textbf{0.941} & 7.44  & 0.75 & \underline{\textbf{0.90}} \\
 & Gemini 2.5 Pro    & 0.924 & 6.51  & 0.81 & 0.80 \\
 & Gemini 2.5 Flash  & 0.905 & \underline{\textbf{5.08}}  & 0.74 & 0.65 \\
 & Claude Opus 4     & 0.878 & 13.85 & \textbf{0.86} & 0.84 \\
 & Claude Sonnet 4   & 0.840 & 24.44 & 0.80 & 0.83 \\
 & GPT-4.1           & 0.834 & 48.12 & \textbf{0.86} & 0.88 \\
 & GPT-4.1 Mini      & 0.733 & 58.93 & 0.78 & 0.80 \\
 & GPT-4o            & 0.713 & 37.48 & 0.68 & 0.76 \\
 & LLaMA 3 70B       & 0.244 & 123.33& 0.42 & 0.72 \\
 & Claude 3.5 Sonnet & 0.062 & 124.00& 0.38 & 0.59 \\
 & GPT-4.1 Nano      & 0.000 & 0.00  & 0.13 & 0.60 \\
\midrule
\multirow{12}{*}{\textbf{Workflow}}
 & GPT-4o            & \textbf{0.938} & 7.19  & 0.71 & 0.70 \\
 & GPT-4.1           & 0.934 & 7.48  & 0.82 & 0.83 \\
 & LLaMA 3 70B       & 0.932 & 8.64  & 0.51 & 0.71 \\
 & Gemini 2.5 Pro    & 0.929 & 6.90  & 0.69 & 0.66 \\
 & Gemini 2.5 Flash  & 0.928 & \textbf{6.60}  & 0.50 & 0.60 \\
 & GPT-o3            & 0.925 & 7.59  & \underline{\textbf{0.88}} & \textbf{0.84} \\
 & GPT-4.1 Mini      & 0.922 & 7.72  & 0.72 & 0.81 \\
 & Claude 3.5 Sonnet & 0.917 & 7.83  & 0.82 & 0.82 \\
 & Claude Opus 4     & 0.917 & 7.92  & 0.71 & 0.78 \\
 & Claude Sonnet 4   & 0.917 & 8.00  & 0.77 & 0.81 \\
 & GPT-4.1 Nano      & 0.904 & 8.96  & 0.64 & 0.66 \\
\midrule
\multirow{13}{*}{\textbf{Hybrid}}
 & GPT-4o            & \underline{\textbf{0.943}} & 7.15  & 0.77 & 0.76 \\
 & GPT-4.1           & 0.941 & 7.29  & \textbf{0.82} & 0.80 \\
 & LLaMA 3 70B       & 0.938 & 7.52  & 0.60 & 0.76 \\
 & Gemini 2.5 Pro    & 0.938 & 6.52  & 0.74 & 0.75 \\
 & Claude 3.5 Sonnet & 0.935 & 7.32  & 0.81 & 0.83 \\
 & GPT-o3            & 0.932 & 9.33  & \textbf{0.82} & \textbf{0.87} \\
 & Gemini 2.5 Flash  & 0.932 & \textbf{6.15}  & 0.59 & 0.66 \\
 & GPT-4.1 Mini      & 0.929 & 7.26  & 0.81 & 0.79 \\
 & Claude Opus 4     & 0.920 &10.46  & \textbf{0.82} & 0.81 \\
 & Claude Sonnet 4   & 0.912 &10.27  & \textbf{0.82} & 0.81 \\
 & GPT-4.1 Nano      & 0.869 & 6.94  & 0.48 & 0.63 \\

\bottomrule
\end{tabular}
\caption{LLM performance by mode, sorted within each mode by Avg Removal Ratio 3T. 
Avg Removal Ratio 3T measures the percentage of tools removed within a rolling 3-turn window after addition, indicating how well a model manages short-term memory over time. 
Avg Residual 3T captures the average number of tools remaining after three turns, reflecting tool accumulation behavior.}
\label{tab:main_results}
\end{table*}

In Algorithm~\ref{alg:hybrid}, the messages and existing tools are initialized as they were in Autonomous Agent and Workflow Mode. Then, the prune LLM call removes irrelevant tools based on the current message state, which exclude the SearchTools function. Finally, the LLM agent enters the while loop, dynamically querying the tool knowledge base for available tools or MCP servers, and executing them to solve the user query.

\paragraph{Recommendations for Hybrid Mode}
\begin{itemize}[nosep]
  \item The added SearchTool enables re-querying the knowledge base if the retrieved tools are not sufficient in answering the user query.
  \item The LLM agent is blinded or abstracted from any removal of tools. It could add too many tools and trigger a tool limit and cause complexity in the workflow.
  \item As LLM models or prompting techniques advance, Hybrid Mode can transition into Autonomous Agent mode.
\end{itemize}

\subsection{Dataset}
We use 5,000 tools or MCP servers from ScaleMCP and sample 100 instances for the 100 multi-turn tool-use dialogue. The sampling method is stratified sample, based on the number of tool calls per turn. ScaleMCP has an average of 5.0 tool calls per turn \cite{lumer2025scalemcpdynamicautosynchronizingmodel}.

\section{Evaluations}\label{sec:evaluations}
\subsection{Experiment Settings}
For all experiments, we evaluate 13 LLM models, ranging from closed and open source models. For the tool knowledge base, the embedding model used is Azure OpenAI's text-embedding-ada-002, since ScaleMCP \cite{lumer2025scalemcpdynamicautosynchronizingmodel} already extensively tested 5 embedding models, 3 reranking models, and 5 retriever types-and found minimal difference between embedding models. We impose a tool count API error limit and set it at 128 to standardize and account for all LLM models (OpenAI GPT series), even though models such as Gemini 2.5 Pro and Claude Opus 4 can use 256-512 tools in the function calling API request \cite{openaifunctioncalling2024,google_gemini_2025}.

\paragraph{Multi-Turn Conversation.} All experiments consist of 100 sequential user queries, with an average of 5 tool calls per turn. Each query can be related or unrelated to the immediate prior query. 
\begin{equation}
  \label{eq:removal_ratio}
  \text{RemovalRatio}\;=\;
  \frac{\displaystyle\sum_{t=1}^{T} R_{t}}
       {\displaystyle\sum_{t=1}^{T} A_{t}}
  \;\;\le 1,
\end{equation}

\begin{equation}
\label{eq:avg_removal_ratio_3T}
\begin{aligned}
\text{AvgRemovalRatio}_{3T}
  &= \frac{1}{T-2}
     \sum_{t=3}^{T}
     \frac{\sum_{i=t-2}^{t} R_{i}}
          {\sum_{i=t-2}^{t} A_{i}}
\end{aligned}
\end{equation}

\begin{equation}
\label{eq:avg_residual_3T}
\text{AvgResidual}_{+3} =
\frac{1}{K}\sum_{k=1}^{K}
\left( \frac{1}{3}\sum_{i=1}^{3} T_{,p_{k}+i} \right),
\end{equation}

\begin{equation}
\label{eq:tcomp}
\text{TaskCompletion} = \mathrm{Alignment}(\text{Task}, \text{Output})
\end{equation}

\begin{equation}\label{eq:tcorr}
\text{Tool Correctness} =
\frac{\text{Num.\ Correct Calls}}{\text{Total Calls}}
\end{equation}

\paragraph{Metrics.} Across 100 sequential multi-turn conversations, we primarily track the tool count at each turn to observe how the active tools fluctuate over turns. We define three key tool-calling context engineering metrics: the \textit{Removal Ratio}, which measures the proportion of tools successfully removed after being added; the \textit{Average Residual} after a 3-turn grace window, which captures how many tools remain shortly after each peak tool count; and the \textit{Average Removal Ratio 3T} over a 3-turn window (Equation~\ref{eq:avg_removal_ratio_3T}), which smooths removal behavior by averaging removal ratios computed across successive 3-turn windows. For each turn, we use a \textit{Task Completion Score} (Equation~\ref{eq:tcomp}), computed using LLM-as-judge alignment scoring between the agent’s final response and the expected answer in the benchmark \cite{deepeval2025}. We use OpenAI GPT-4o mini as the judge LLM model in all evaluatios. Also, we compute the \textit{Tool Correctness Score} (Equation~\ref{eq:tcorr}), which compares each tool call to the expected tool call \cite{deepeval2025}. 

\begin{figure*}[t]
  \centering
  \includegraphics[width=\textwidth]{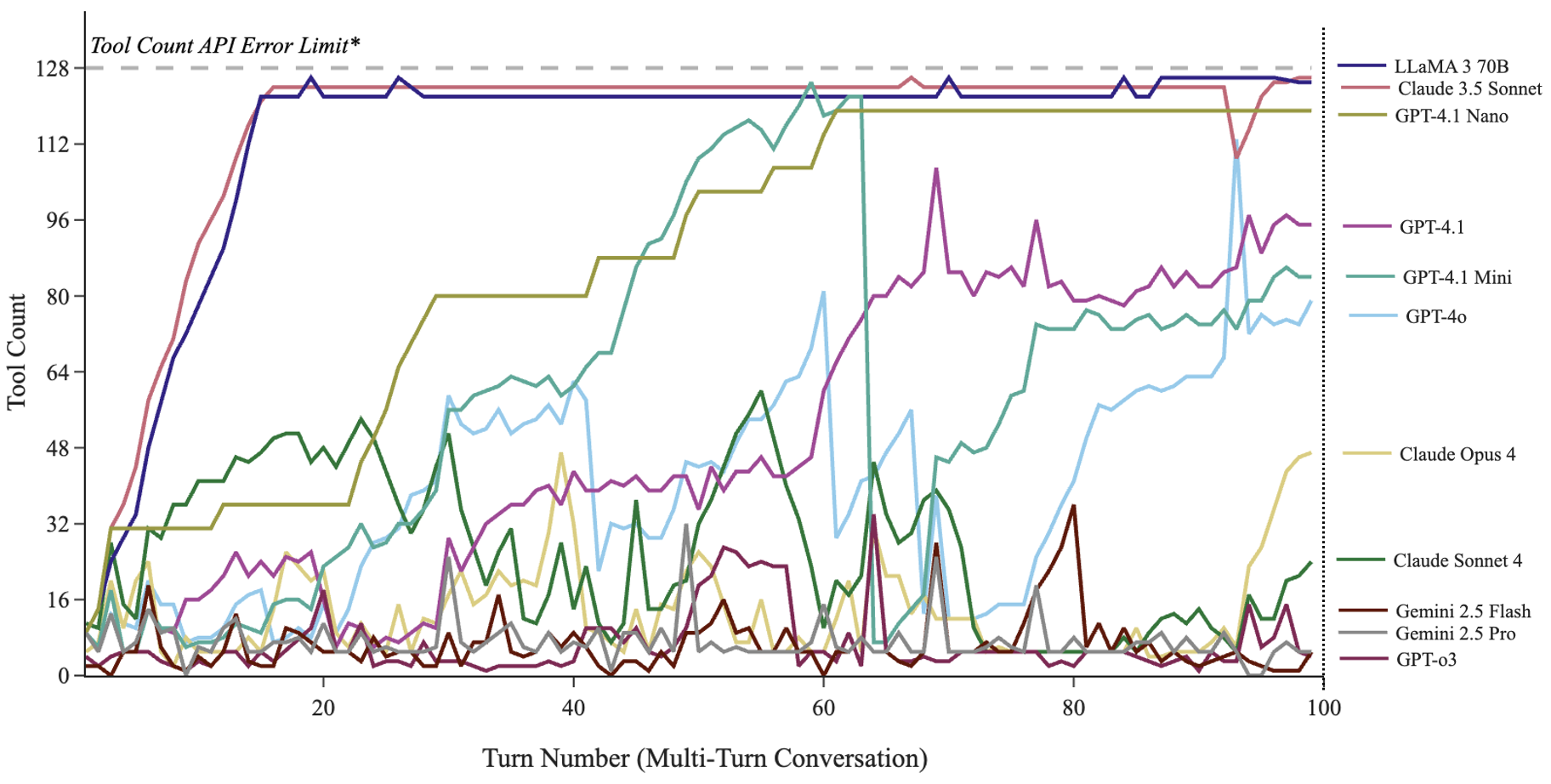} 
    \caption{MemTool Autonomous Agent Mode: Tool Count across 100 multi-turn queries for various LLMs. While some models (e.g., GPT-o3, Gemini 2.5 Pro) maintain stable tool windows and consistently remove irrelevant tools added in preivous turns, others (e.g., GPT-4.1 Nano, Claude 3.5 Sonnet, LLaMA 3 70B) fail to remove unused tools, exceeding the 128-tool API limit. This highlights large variation in agents' ability to manage their short-term tool memory. Note: Tool Count API error Limit* is set at 128 to standardize and account for all LLM models, even though models such as Gemini 2.5 Pro and Claude Opus 4 can use 256-512 tools in the function calling API request.}

  \label{fig:autonmous_graph}
\end{figure*}

\subsection{Autonomous Agent Mode Multi-Turn Experiments}
In autonomous agent mode, we equip an LLM agent with two tools to manage its tool context and track how well it manages its context across 100+ sequential user queries, while answering the user with any new tools it retrieves. 
\subsubsection{Results Analysis}
As seen in Table~\ref{tab:main_results} and Figure~\ref{fig:autonmous_graph}, there is a wide range of performance across LLM models on how they manage their short-term tool memory across 100 sequential multi-turn conversations. The highest performing LLM models, denoted as ones that have high 3-window average removal ratios and task completion, consist of reasoning models such as Gemini 2.5 Pro/Flash, OpenAI GPT-o3, and Clause Opus 4. These models remove roughly 95\% of their total tools across 100 conversations, and 87.8-100\% of their tools on a rolling 3-window period. Visually, these models handle spikes in their tool count effectively, with any increase in tool count followed by a removal of unused tools for subsequent queries (Figure~\ref{fig:autonmous_graph}). This is shown by the high 3-window rolling average removal rates. Furthermore, these models are strong at simultaneously answering the user query, with Task Completion and Tool Correctness metrics averaging between 80-90\%.

On the other hand, smaller and medium sized models, or models that were not post-trained with large reinforcement learning reasoning budgets, did not effectively manage their available tools while answering the user query. These models include OpenAI GPT-4o Mini, GPT-4.1 Mini, Anthropic Claude 3.5 Sonnet, and LlaMa 3 70B. Gemini 2.5 Flash is the only small model that had a rolling average 3-window removal ratio above 90\%. However, the same model struggled to effectively answer the user query. Visually, these models struggle to remove unused tools, and are more focused on adding tools to their context window. As seen in Figure~\ref{fig:autonmous_graph}, LlaMa 3 70B and Claude 3.5 Sonnet climb to the 128 limit within turn 20, and the other models tend to add more tools than they remove over time, causing sporadic spikes and steady climing spikes. 

For the full results of all LLM models evaluated, see Appendix~\ref{app:full_table}, Table~\ref{tab:full_main_results} and Appendix~\ref{app:agentmode}, Figures~\ref{fig:agent-openai-o3} through \ref{fig:agent-llama3}.

\subsubsection{Discussion}
 Medium and small LLM models in MemTool Autonomous Agent Mode struggle to efficiently manage its short-term tool memory while answering the user query. We recommend using reasoning models that have been post-trained with large reinforcement learning budgets, which impact how well models are able to manage its tool context window. The benefits of giving the LLM agent full autonomy over its tool context window consist of re-searching the tool knowledge base for more tools if it cannot find one the first time. While some models are able to remove irrelevant tools, others are too focused on answering the user query and forget to remove tools. Moreoever, the searching and removing mechanism is fully dependent on the system prompt. We found very poor removal rates when the system prompt did not include the current number of tools in its context. To bypass this limitation, we pass a dynamic variable in the system prompt on every query, which is the length of the current tool count. Our results point to a clear distinction in the abilities of certain LLM models when handling short-term dynamic tool memory.

 Common errors for Autonomous Agent Mode entail letting the tool count in the context window reach beyond the set 128 limit. Some models hover around this limit and do not comprehend that it can remove 100 irrelevant tools at a time. Furthermore, we found that the system prompt, when it did not include the current tool count, caused the LLM agent to not remove tools. This indicates an inability for the LLM to understand the count of available tools it has, when interacting with the tools or function calling parameter in the API version of LLM models \cite{openaifunctioncalling2024,google2025}. 
\subsection{Workflow Mode Multi-Turn Experiments}
In Workflow Mode, LLMs operate under a deterministic short-term memory framework where tool context is centrally managed. Instead of granting the agent autonomy, the system prunes irrelevant tools and adds new ones through separate LLM calls, enforcing memory optimization before the agent answers each query.

\subsubsection{Results Analysis}
\begin{figure}[t]
  \centering
  \includegraphics[width=\linewidth]{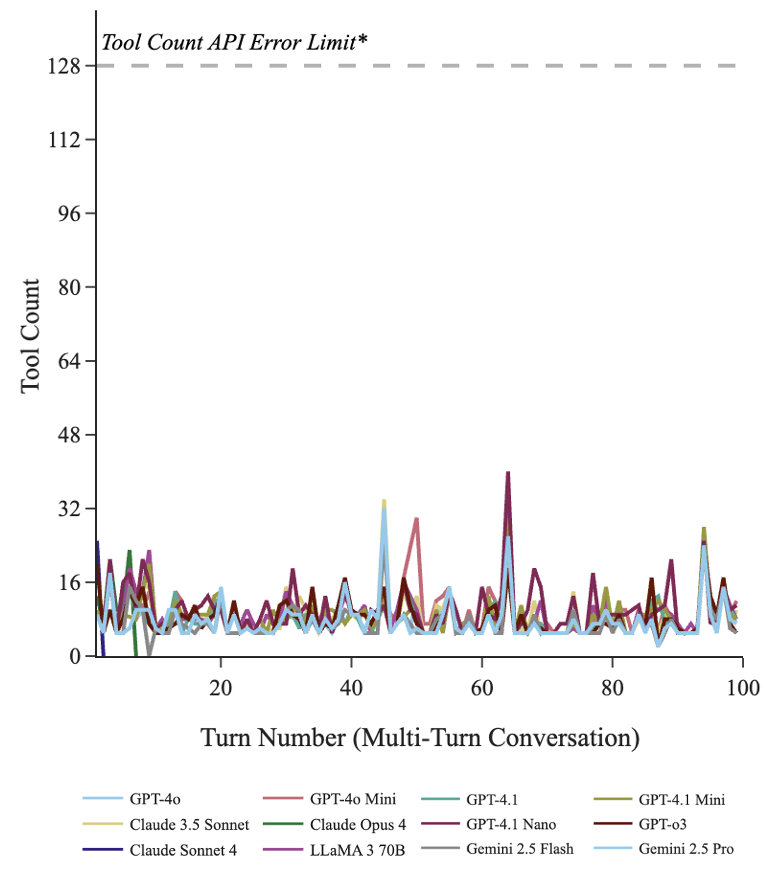}
  \caption{MemTool Workflow Mode: Tool Count across 100 multi-turn queries for various LLMs. All models maintain tool counts below the 128-tool API threshold.}
  \label{fig:workflow_graph}
\end{figure}

As seen in Figure~\ref{fig:workflow_graph} and Table~\ref{tab:main_results}, Workflow Mode yielded consistently high performance across nearly all tested LLM models. The average 3-window removal ratio remained above 90\% for all models, with GPT-4o, GPT-o3, and Gemini 2.5 Pro achieving top-tier performance in both removal and task metrics. Notably, all models avoided excessive tool accumulation across the 100-turn conversation window, staying well below the API error threshold of 128 tools.

Tool Correctness and Task Completion varied more than removal efficiency. GPT-o3 achieved the highest Tool Correctness score (88\%), followed closely by GPT-4.1 and Claude 3.5 Sonnet (both 82\%). In Task Completion, GPT-o3 led at 84\%, followed by GPT-4.1 and Claude 3.5 Sonnet again led with 83\% and 82\%, respectively. Lower-performing models in these categories included GPT-4.1 Nano and Gemini 2.5 Flash, which, while excellent at tool removal, struggled with reasoning over tools to answer the user question.

\subsubsection{Discussion}
The deterministic nature of Workflow Mode proves highly effective for short-term memory management. Since all LLM models are guided by the same two-step pipeline (prune-then-search), they consistently maintain a lean tool context and avoid tool accumulation issues seen in Autonomous Agent Mode. This removes the burden of memory optimization from the model, allowing even smaller models to perform well.

We recommend using Workflow Mode when cost efficiency and reliability are priorities. Given that nearly all models in this mode achieve comparable Removal Ratio and AvgRemovalRatio3T, the decision of which LLM to use can hinge on cost and performance in downstream tasks. For instance, if high Tool Correctness is critical, GPT-o3 or GPT-4.1 are preferable; if Task Completion is more important, Claude 3.5 Sonnet and GPT-4.1 stand out. In production settings, a low-cost model like GPT-4.1 Mini or Gemini 2.5 Flash can serve as the memory controller, while a more capable model handles the user-facing task.

\subsection{Hybrid Mode Multi-Turn Experiments}
Hybrid Mode blends agentic flexibility with structured control by allowing the LLM to add tools via \texttt{SearchTools}, while delegating tool removal to a separate pruning step. This balances dynamic retrieval with stable memory management across multi-turn conversations.

\subsubsection{Results Analysis}
\begin{figure}[t]
  \centering
  \includegraphics[width=\linewidth]{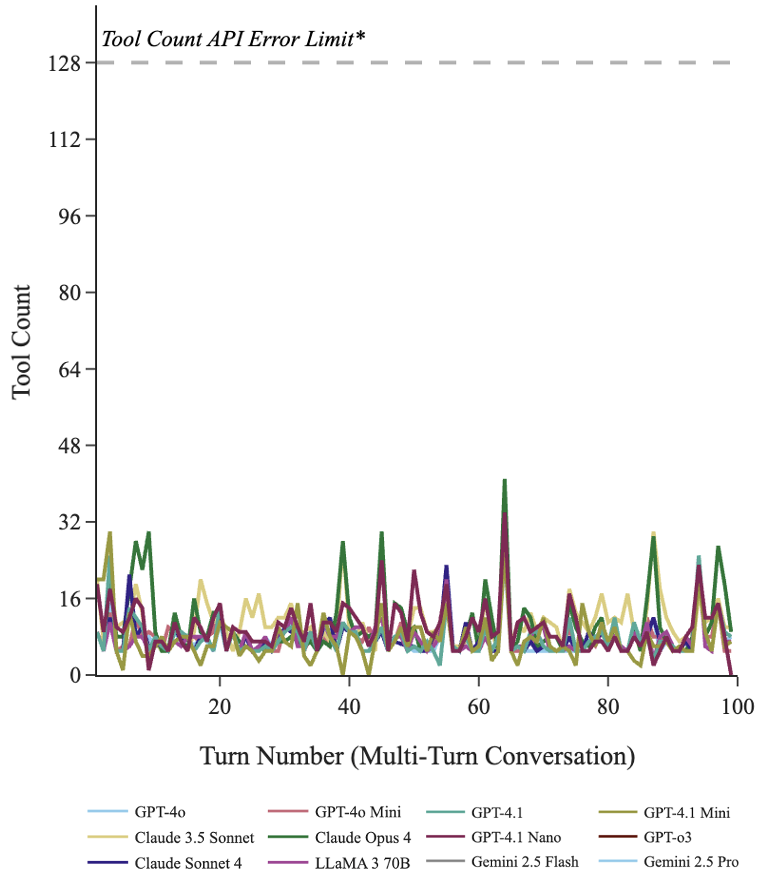}
  \caption{MemTool Hybrid Mode: Tool Count across 100 multi-turn queries for various LLMs. Most models stay stable and far below the 128-tool limit.}
  \label{fig:hybrid_graph}
\end{figure}

As shown in Figure~\ref{fig:hybrid_graph} and Table~\ref{tab:main_results}, Hybrid Mode demonstrates consistently strong performance across LLMs in short-term memory management. Nearly all models achieve average 3-window removal ratios above 90\%, with OpenAI o3 achieving a perfect 100\% RemovalRatio and 93\% AvgRemovalRatio3T. Task Completion and Tool Correctness scores are also generally high, led by Claude 3.7 Sonnet (88\% and 83\%), OpenAI o3 (87\% and 82\%), and GPT-4.1 (80\% and 82\%).

Claude Sonnet 4 and Claude Opus 4 perform well in correctness and task alignment, although they show slightly elevated average residuals after 3 turns (above 10). Gemini 2.5 Flash and GPT-4.1 Nano are the only models that fall below a 70\% Task Completion threshold, consistent with their lower correctness scores (66\% and 48\% respectively). Notably, all models remain well below the 128-tool API limit, indicating successful memory regulation even in the absence of agent-led pruning.

\subsubsection{Discussion}
Hybrid Mode successfully balances structure and autonomy, enabling agents to correct and expand their toolset when needed, while avoiding tool bloat via deterministic pruning. This decoupled design helps stabilize tool count while maintaining agent adaptability.

OpenAI o3 emerges as one of the top-performing models overall, achieving perfect removal and the second-highest Task Completion (87\%). Claude 3.7 Sonnet leads in task alignment (88\%), followed closely by GPT-4.1 and Claude 3.5 Sonnet. While smaller models like Gemini 2.5 Flash and GPT-4.1 Nano exhibit lower downstream performance, they still maintain high removal ratios, suggesting Hybrid Mode allows efficient memory use even for less capable models.

We recommend selecting a model for Hybrid Mode based on task-level metrics such as Tool Correctness and Task Completion. When removal behavior is comparable across models (as seen in Table~\ref{tab:main_results}), performance in reasoning and retrieval becomes the differentiating factor.

\section{Conclusion}

As Large Language Model (LLM) agents demonstrate significant capabilities in dynamically searching and incorporating relevant tools or MCP servers, fixed context windows limit their effectiveness in multi-turn interactions requiring repeated, independent tool usage. We introduced MemTool, a short-term memory framework enabling LLM agents to dynamically manage their context window of tools or MCP servers across multi-turn conversations. MemTool provides three agentic architectures with varying autonomy levels: 1) Autonomous Agent Mode (full tool management autonomy), 2) Workflow Mode (deterministic control without autonomy), and 3) Hybrid Mode (combining autonomous and deterministic control). Evaluating each MemTool mode across 13+ LLM models on the ScaleMCP benchmark (5,000 MCP servers), we found that reasoning LLMs in Autonomous Agent Mode achieved high tool-removal efficiency (90–94\% over a 3-window average), whereas medium-sized models demonstrated significantly lower efficiency (0–60\%). Workflow and Hybrid modes consistently managed tool removal effectively, while Autonomous and Hybrid modes excelled in adaptive tool addition and task completion. MemTool pushes the needle forward in short-term memory for tool-using LLM agents, providing evidence that LLM agents can effectively manage their context windows of dynamic tools. Looking forward, MemTool can be paired with long-term memory advancements in tool-using LLM agents.

\section*{Limitations}
Although we demonstrate MemTool as reliable for dynamic tool management within short-term memory, it does have several limitations. MemTool Autonomous Agent Mode struggles with reliably removing tools, especially when non-reasoning models are used; this limitation can be addressed by using the MemTool Workflow Mode, which is deterministic. However, MemTool Workflow Mode itself constrains the agent’s capacity to loop back and explore additional tools; thus,  MemTool Hybrid Mode becomes beneficial, balancing deterministic structure with agent autonomy. Nevertheless,  MemTool Hybrid Mode encounters limitations if the agent explores too many tools and breaches predefined limits; one feasible mitigation strategy involves invoking a dedicated pruning LLM, though this approach can inadvertently remove essential context, suggesting a potential need to temporarily revert to  MemTool Autonomous Agent Mode. Additionally, both the Workflow and Hybrid modes require careful selection of the appropriate LLM. We address this limitation by providing Tool Correctness and Task Completion metrics for each LLM, allowing for an informed selection or mixing of LLMs based on their respective strengths, such as proficiency in tool removal and searching versus actual tool usage. Ultimately, selecting the appropriate MemTool mode aligned with the specific LLM model and task context can effectively mitigate these limitations.


\bibliography{acl_latex}

\appendix
\section*{Appendix}

\section{Full LLM Evaluation Table by Mode}\label{app:full_table}
We show the complete quantitative results across all models and agent modes in Appendix~\ref{app:full_table}, Table~\ref{tab:full_main_results}.

\begin{table*}[t]
\centering
\footnotesize
\begin{tabular}{llrrrr}
\toprule
\textbf{Mode} & \textbf{LLM} & \textbf{Avg Removal Ratio 3T} & \textbf{Avg Residual 3T} & \textbf{Tool Correctness} & \textbf{Task Completion} \\
\midrule
\multirow{13}{*}{\shortstack{\textbf{Autonomous} \\ \textbf{Agent}}}
 & GPT-o3            & \textbf{0.941} & 7.44  & 0.75 & \underline{\textbf{0.90}} \\
 & Gemini 2.5 Pro    & 0.924 & 6.51  & 0.81 & 0.80 \\
 & Claude 3.7 Sonnet & 0.905 & 14.25 & \textbf{0.86} & 0.89 \\
 & Gemini 2.5 Flash  & 0.905 & \underline{\textbf{5.08}}  & 0.74 & 0.65 \\
 & Claude Opus 4     & 0.878 & 13.85 & \textbf{0.86} & 0.84 \\
 & Claude Sonnet 4   & 0.840 & 24.44 & 0.80 & 0.83 \\
 & GPT-4.1           & 0.834 & 48.12 & \textbf{0.86} & 0.88 \\
 & GPT-4.1 Mini      & 0.733 & 58.93 & 0.78 & 0.80 \\
 & GPT-4o            & 0.713 & 37.48 & 0.68 & 0.76 \\
 & GPT-4o Mini       & 0.449 & 121.06& 0.78 & 0.88 \\
 & LLaMA 3 70B       & 0.244 & 123.33& 0.42 & 0.72 \\
 & Claude 3.5 Sonnet & 0.062 & 124.00& 0.38 & 0.59 \\
 & GPT-4.1 Nano      & 0.000 & 0.00  & 0.13 & 0.60 \\
\midrule
\multirow{12}{*}{\textbf{Workflow}}
 & GPT-4o            & \textbf{0.938} & 7.19  & 0.71 & 0.70 \\
 & GPT-4.1           & 0.934 & 7.48  & 0.82 & 0.83 \\
 & LLaMA 3 70B       & 0.932 & 8.64  & 0.51 & 0.71 \\
 & Gemini 2.5 Pro    & 0.929 & 6.90  & 0.69 & 0.66 \\
 & Gemini 2.5 Flash  & 0.928 & \textbf{6.60}  & 0.50 & 0.60 \\
 & GPT-o3            & 0.925 & 7.59  & \underline{\textbf{0.88}} & \textbf{0.84} \\
 & GPT-4.1 Mini      & 0.922 & 7.72  & 0.72 & 0.81 \\
 & Claude 3.5 Sonnet & 0.917 & 7.83  & 0.82 & 0.82 \\
 & Claude Opus 4     & 0.917 & 7.92  & 0.71 & 0.78 \\
 & Claude Sonnet 4   & 0.917 & 8.00  & 0.77 & 0.81 \\
 & GPT-4o Mini       & 0.916 & 8.43  & 0.74 & 0.72 \\
 & GPT-4.1 Nano      & 0.904 & 8.96  & 0.64 & 0.66 \\
\midrule
\multirow{13}{*}{\textbf{Hybrid}}
 & GPT-4o            & \underline{\textbf{0.943}} & 7.15  & 0.77 & 0.76 \\
 & GPT-4.1           & 0.941 & 7.29  & 0.82 & 0.80 \\
 & LLaMA 3 70B       & 0.938 & 7.52  & 0.60 & 0.76 \\
 & Gemini 2.5 Pro    & 0.938 & 6.52  & 0.74 & 0.75 \\
 & Claude 3.5 Sonnet & 0.935 & 7.32  & 0.81 & 0.83 \\
 & GPT-4o Mini       & 0.934 & 7.77  & 0.81 & 0.80 \\
 & GPT-o3            & 0.932 & 9.33  & 0.82 & 0.87 \\
 & Gemini 2.5 Flash  & 0.932 & \textbf{6.15}  & 0.59 & 0.66 \\
 & GPT-4.1 Mini      & 0.929 & 7.26  & 0.81 & 0.79 \\
 & Claude 3.7 Sonnet & 0.921 & 7.97  & \textbf{0.83} & \textbf{0.88} \\
 & Claude Opus 4     & 0.920 &10.46  & 0.82 & 0.81 \\
 & Claude Sonnet 4   & 0.912 &10.27  & 0.82 & 0.81 \\
 & GPT-4.1 Nano      & 0.869 & 6.94  & 0.48 & 0.63 \\

\bottomrule
\end{tabular}
\caption{LLM performance by mode, sorted within each mode by Avg Removal Ratio 3T. 
Avg Removal Ratio 3T measures the percentage of tools removed within a rolling 3-turn window after addition, indicating how well a model manages short-term memory over time. 
Avg Residual 3T captures the average number of tools remaining after three turns, reflecting tool accumulation behavior.}
\label{tab:full_main_results}
\end{table*}

\section{Prompts}\label{app:prompts}
\begin{figure*}[t]
  \centering
  \includegraphics[width=\textwidth]{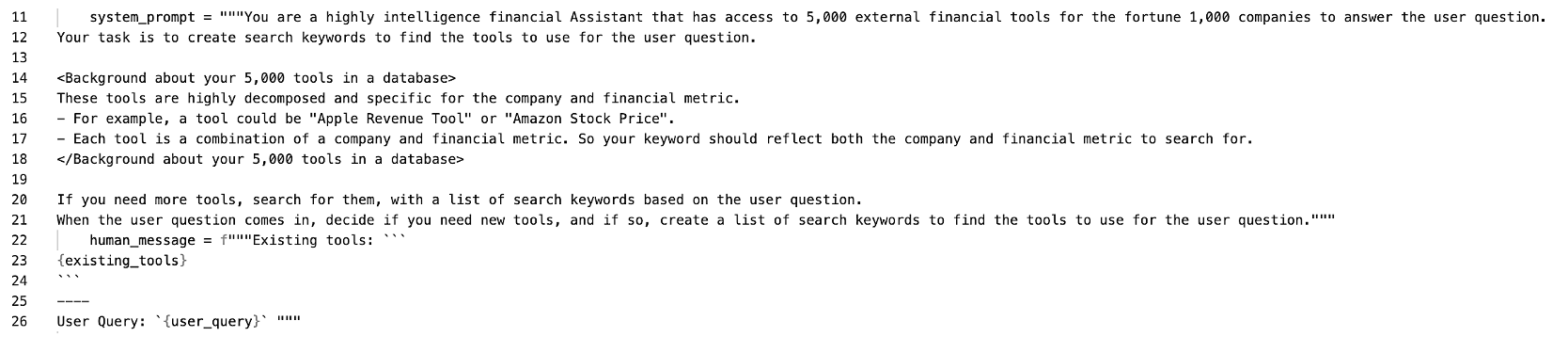}
  \caption{Search Tools prompt for Workflow Mode LLM call to add more tools by generating search keywords based on the user question.} 
  \label{fig:search_prompt}
\end{figure*}

\begin{figure*}[t]
  \centering
  \includegraphics[width=\textwidth]{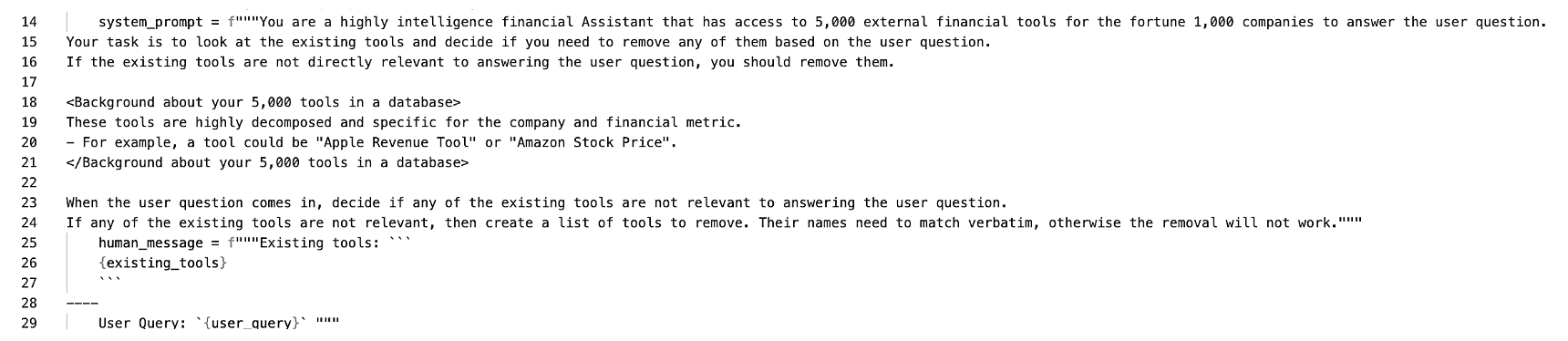}
  \caption{Remove Tools prompt for Workflow Mode and Hybrid Mode LLM call to remove irrelevant tools by responding with the exact tool names based on the user question.} 
  \label{fig:agent_remove_prompt}
\end{figure*}

We include the full prompt templates used by the LLMs in each mode to manage tool search, addition, and removal behavior, as shown in Appendix~\ref{app:prompts}, Figures~\ref{fig:agent_system_prompt}, \ref{fig:search_prompt}, and \ref{fig:agent_remove_prompt}.

\begin{figure*}[t]
  \centering
  \includegraphics[width=\textwidth]{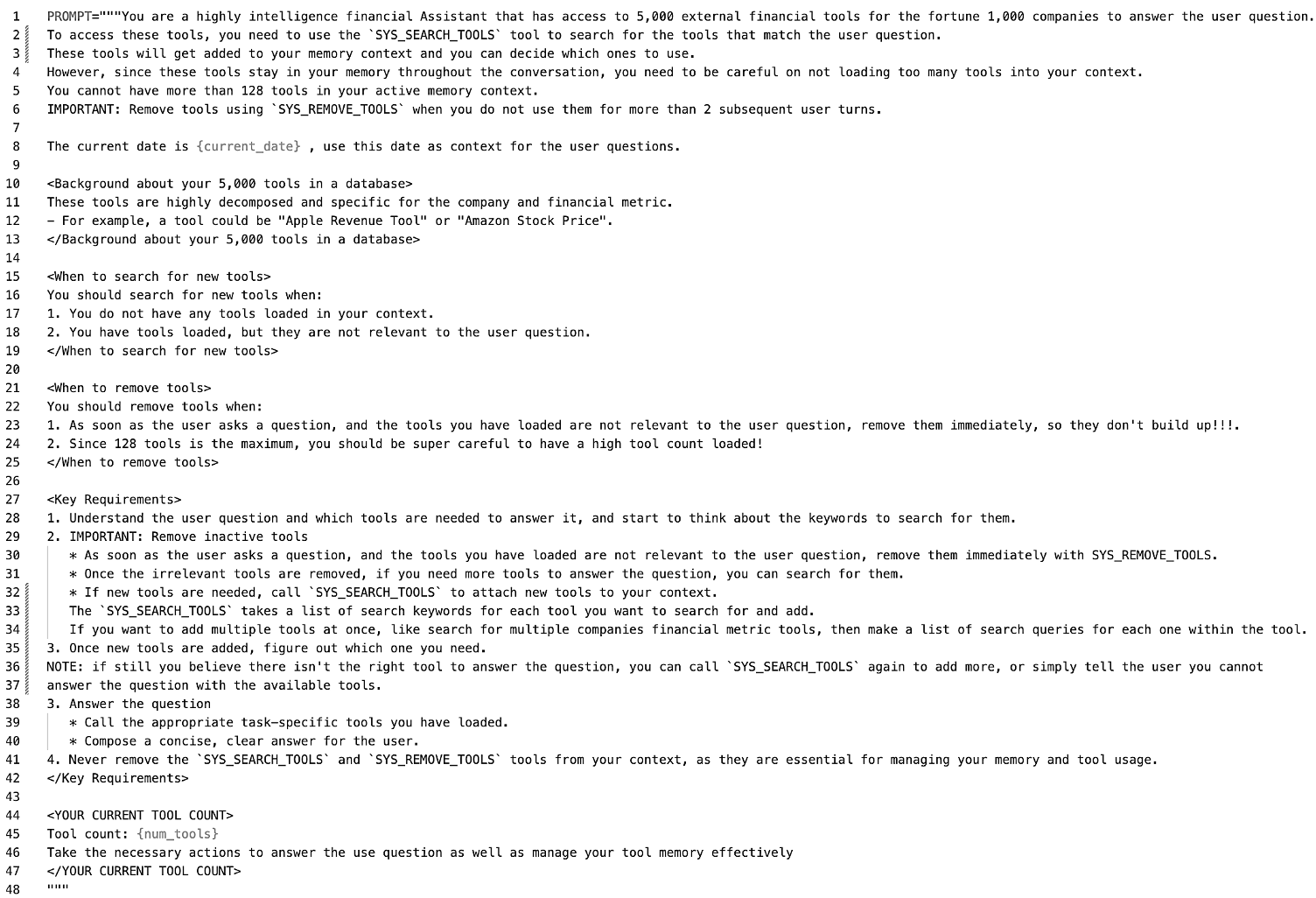}
  \caption{Autonomous Agent System Prompt, with the SearchTool (SYS\_SEARCH\_TOOL) and RemoveTool (SYS\_REMOVE\_TOOL)} 
  \label{fig:agent_system_prompt}
\end{figure*}

\section{Autonomous Agent Mode — Tool Count Graphs}\label{app:agentmode}

This section presents individual tool count trajectories for each LLM evaluated under MemTool's Autonomous Agent Mode across 100 multi-turn conversations. These graphs visualize each model's ability to regulate its short-term memory window.

\begin{figure*}[t]
  \centering
  \includegraphics[width=0.8\textwidth]{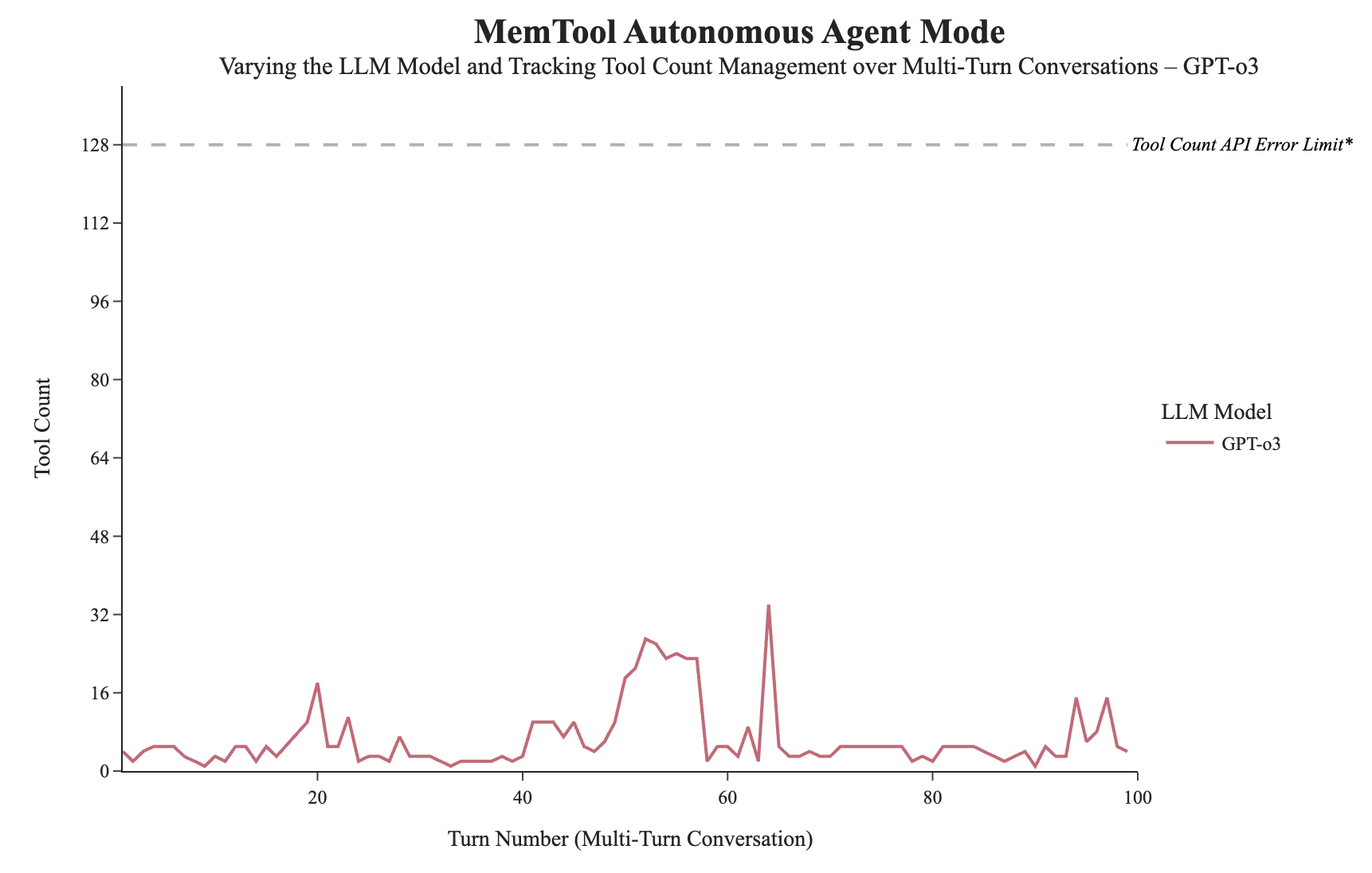}
  \caption{OpenAI o3 — Tool Count over Time (Autonomous Agent Mode)}
  \label{fig:agent-openai-o3}
\end{figure*}

\begin{figure*}[t]
  \centering
  \includegraphics[width=0.8\textwidth]{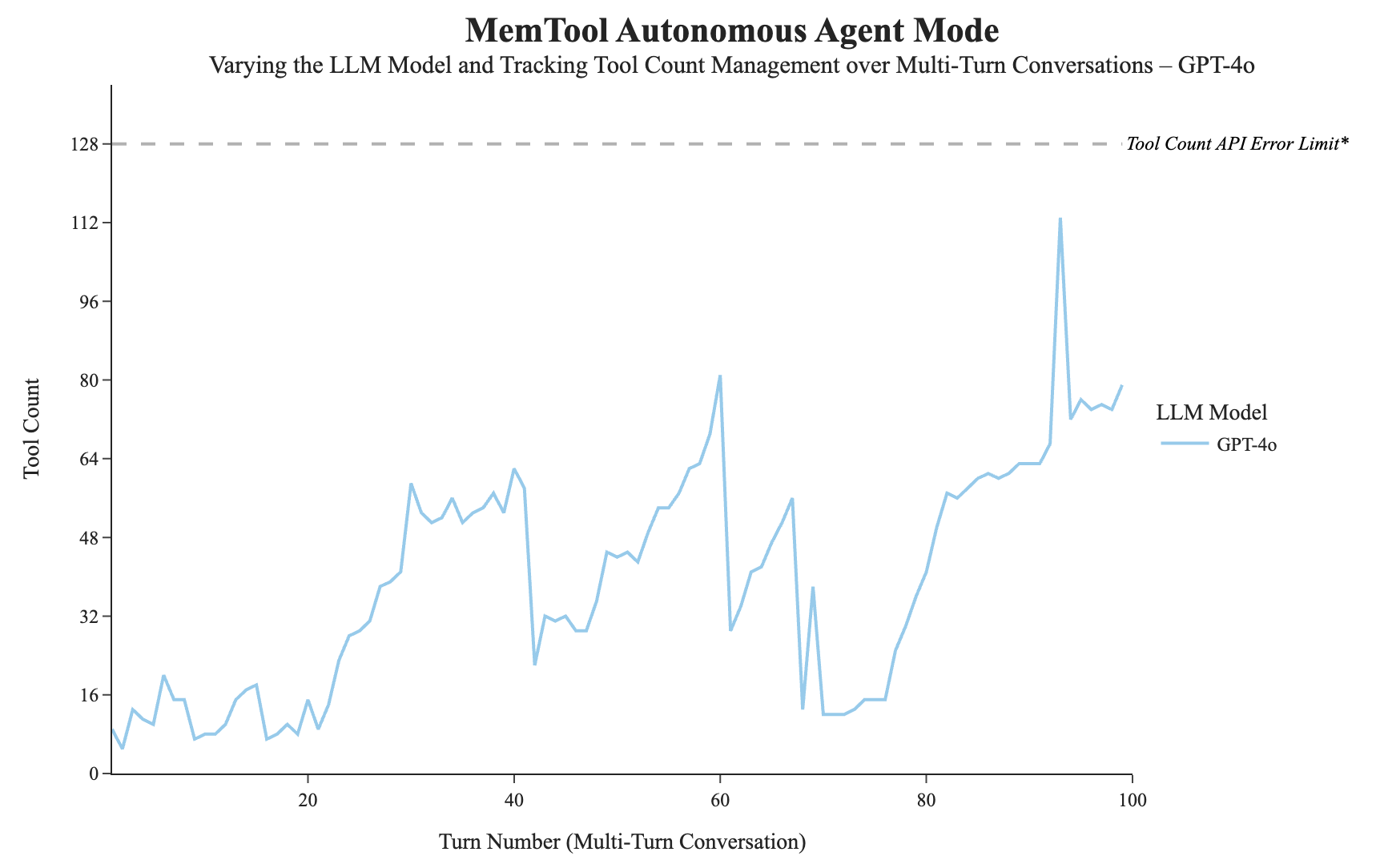}
  \caption{GPT-4o — Tool Count over Time (Autonomous Agent Mode)}
  \label{fig:agent-gpt4o}
\end{figure*}

\begin{figure*}[t]
  \centering
  \includegraphics[width=0.8\textwidth]{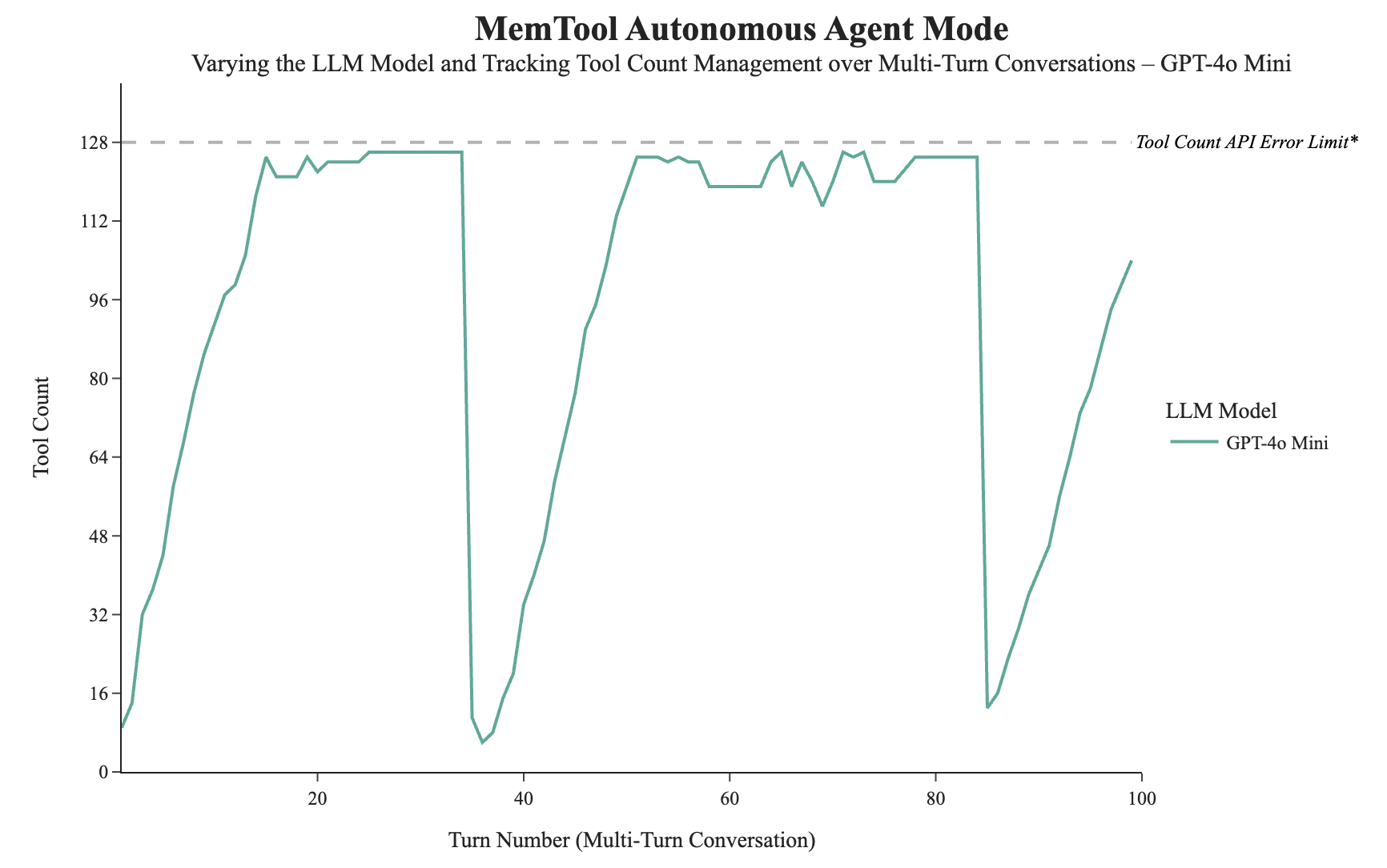}
  \caption{GPT-4o Mini — Tool Count over Time (Autonomous Agent Mode)}
  \label{fig:agent-gpt4o-mini}
\end{figure*}

\begin{figure*}[t]
  \centering
  \includegraphics[width=0.8\textwidth]{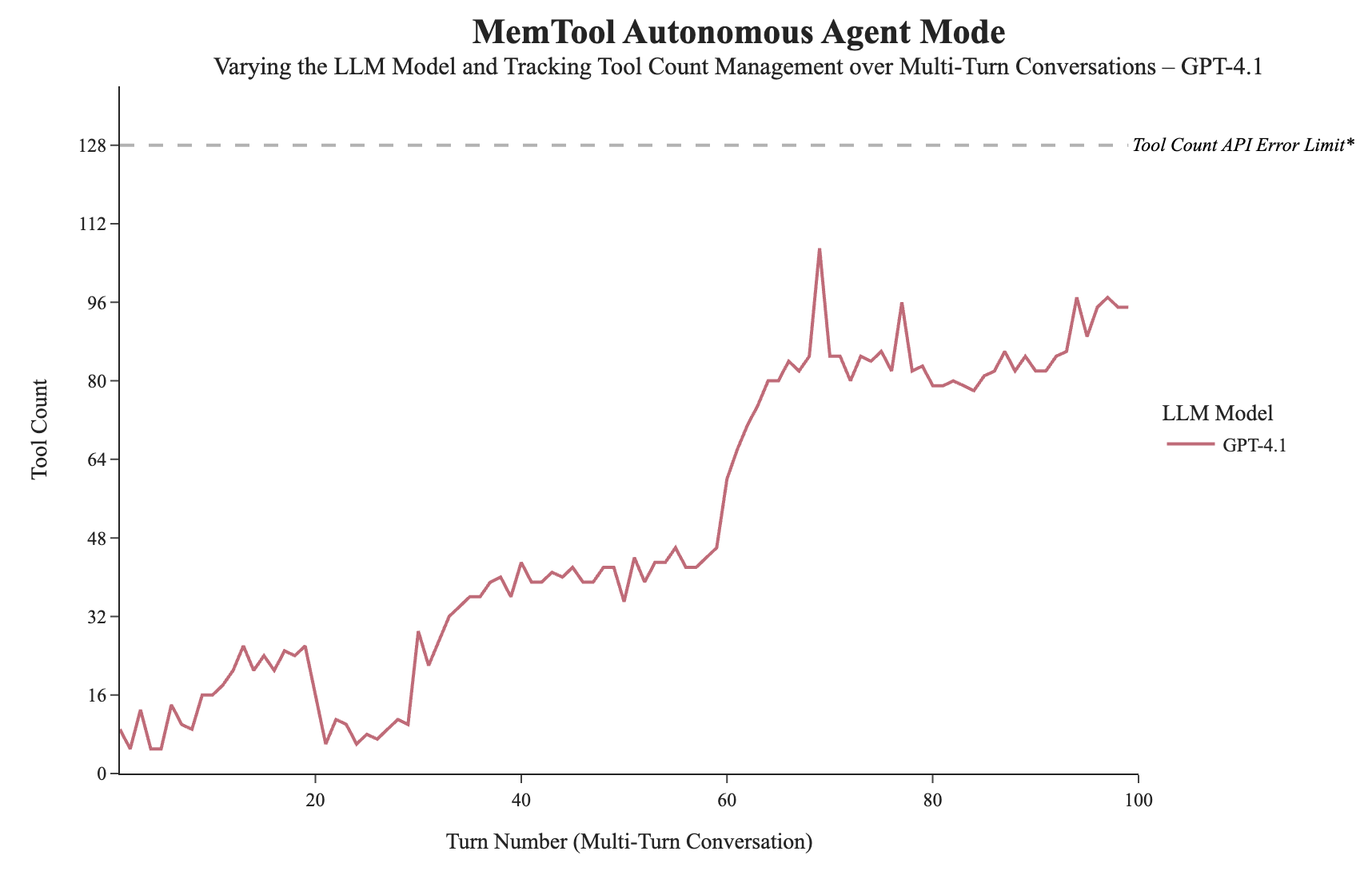}
  \caption{GPT-4.1 — Tool Count over Time (Autonomous Agent Mode)}
  \label{fig:agent-gpt41}
\end{figure*}

\begin{figure*}[t]
  \centering
  \includegraphics[width=0.8\textwidth]{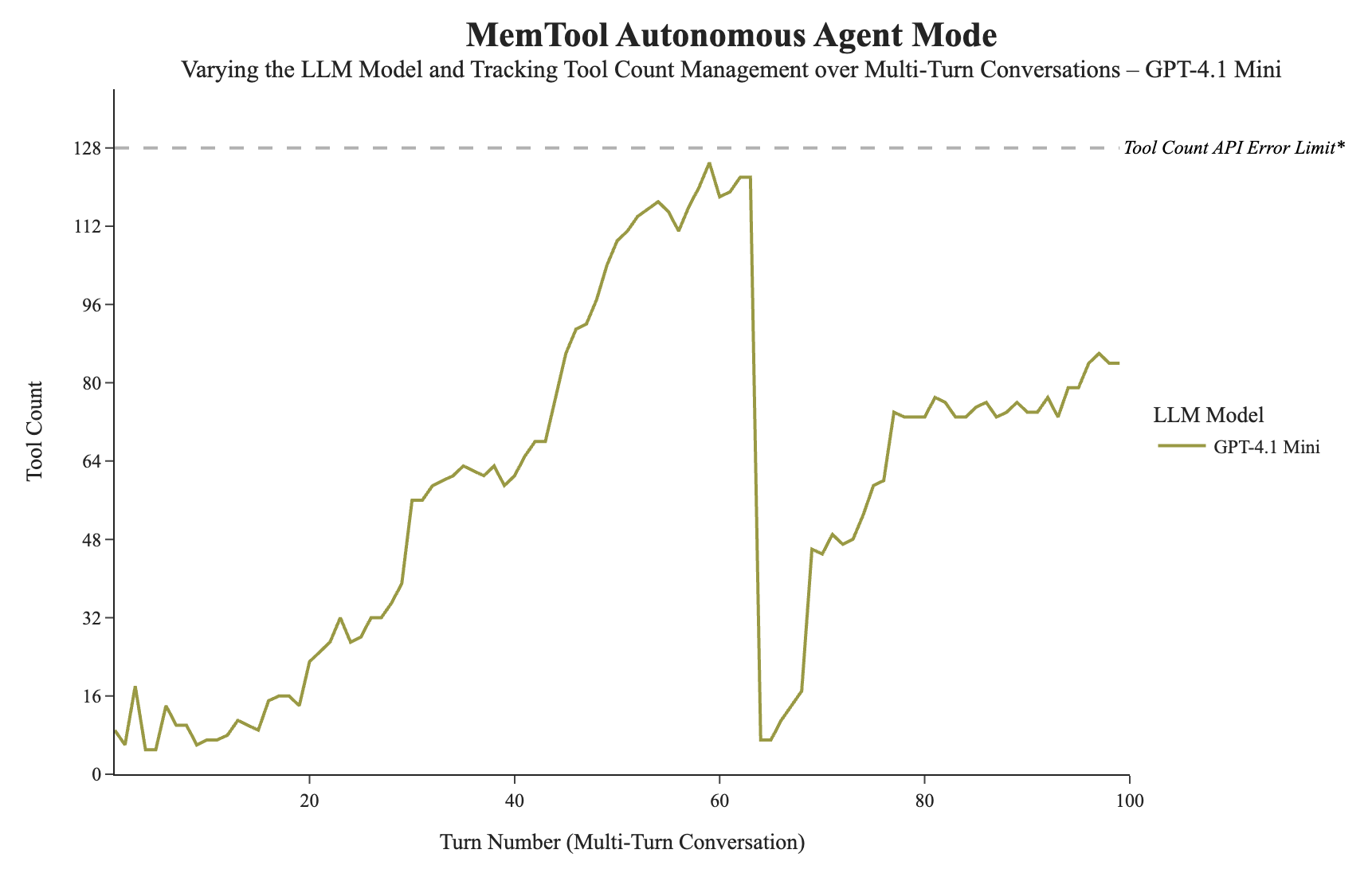}
  \caption{GPT-4.1 Mini — Tool Count over Time (Autonomous Agent Mode)}
  \label{fig:agent-gpt41-mini}
\end{figure*}

\begin{figure*}[t]
  \centering
  \includegraphics[width=0.8\textwidth]{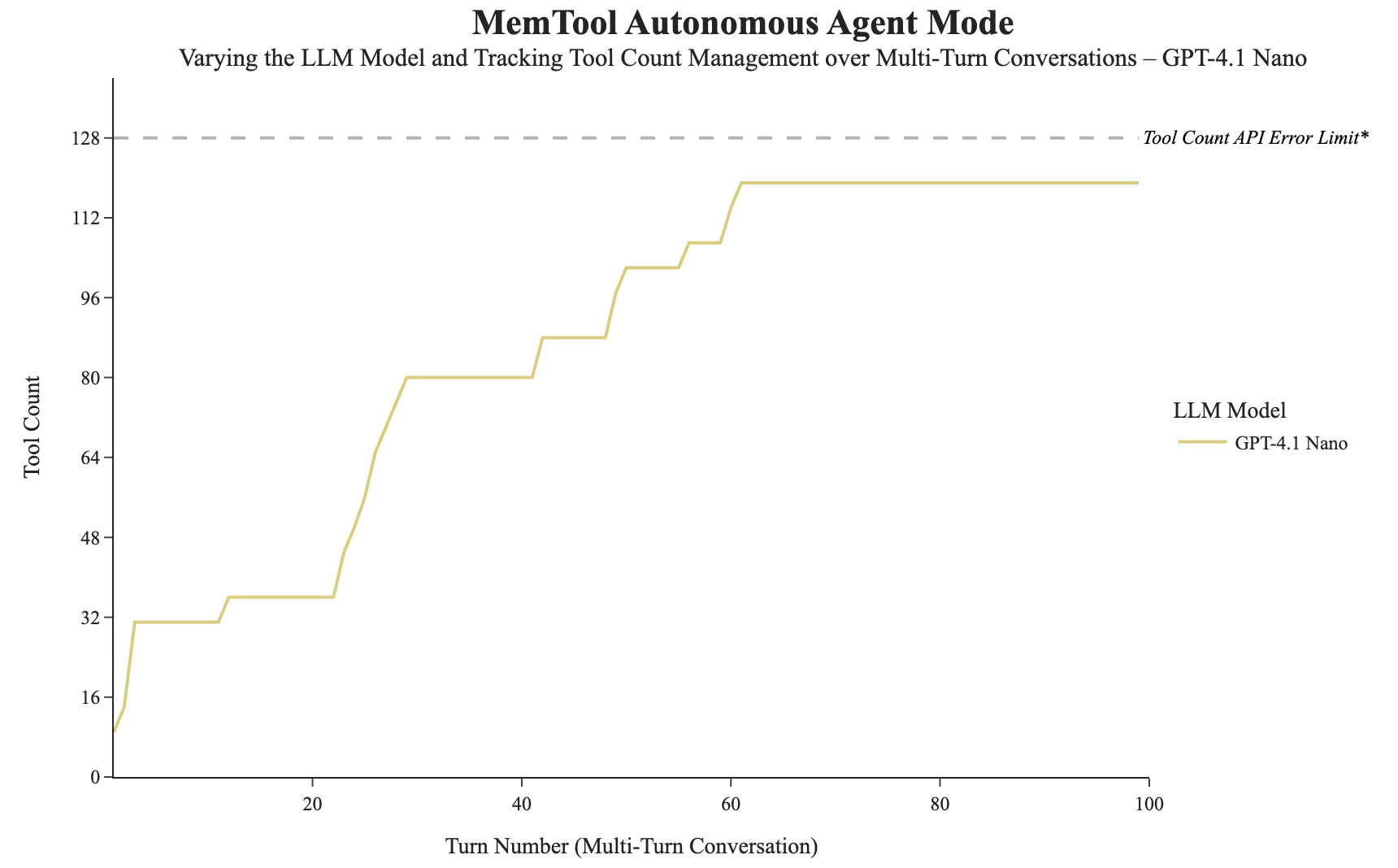}
  \caption{GPT-4.1 Nano — Tool Count over Time (Autonomous Agent Mode)}
  \label{fig:agent-gpt41-nano}
\end{figure*}

\begin{figure*}[t]
  \centering
  \includegraphics[width=0.8\textwidth]{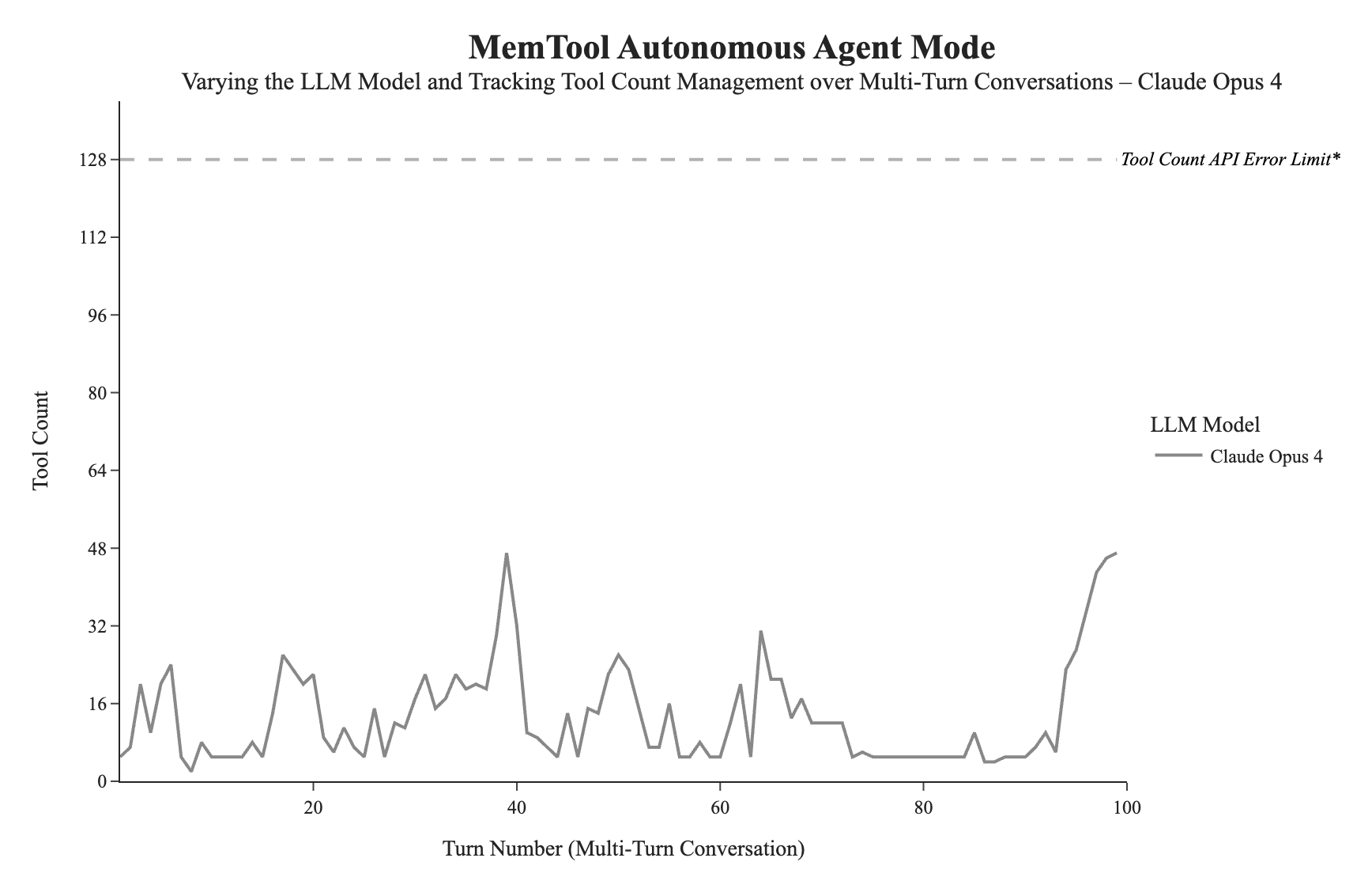}
  \caption{Claude Opus 4 — Tool Count over Time (Autonomous Agent Mode)}
  \label{fig:agent-claude-opus4}
\end{figure*}

\begin{figure*}[t]
  \centering
  \includegraphics[width=0.8\textwidth]{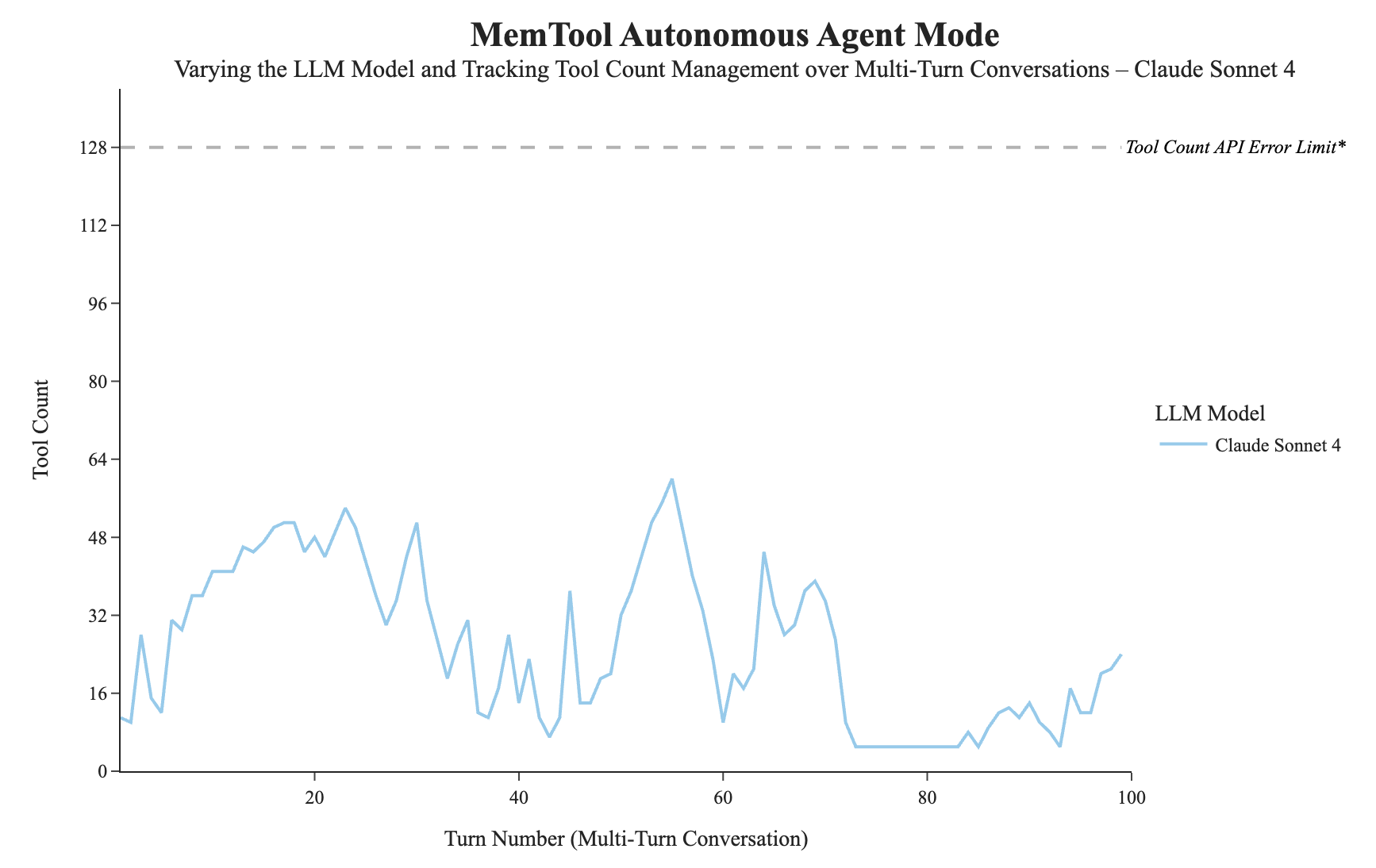}
  \caption{Claude Sonnet 4 — Tool Count over Time (Autonomous Agent Mode)}
  \label{fig:agent-claude-sonnet4}
\end{figure*}

\begin{figure*}[t]
  \centering
  \includegraphics[width=0.8\textwidth]{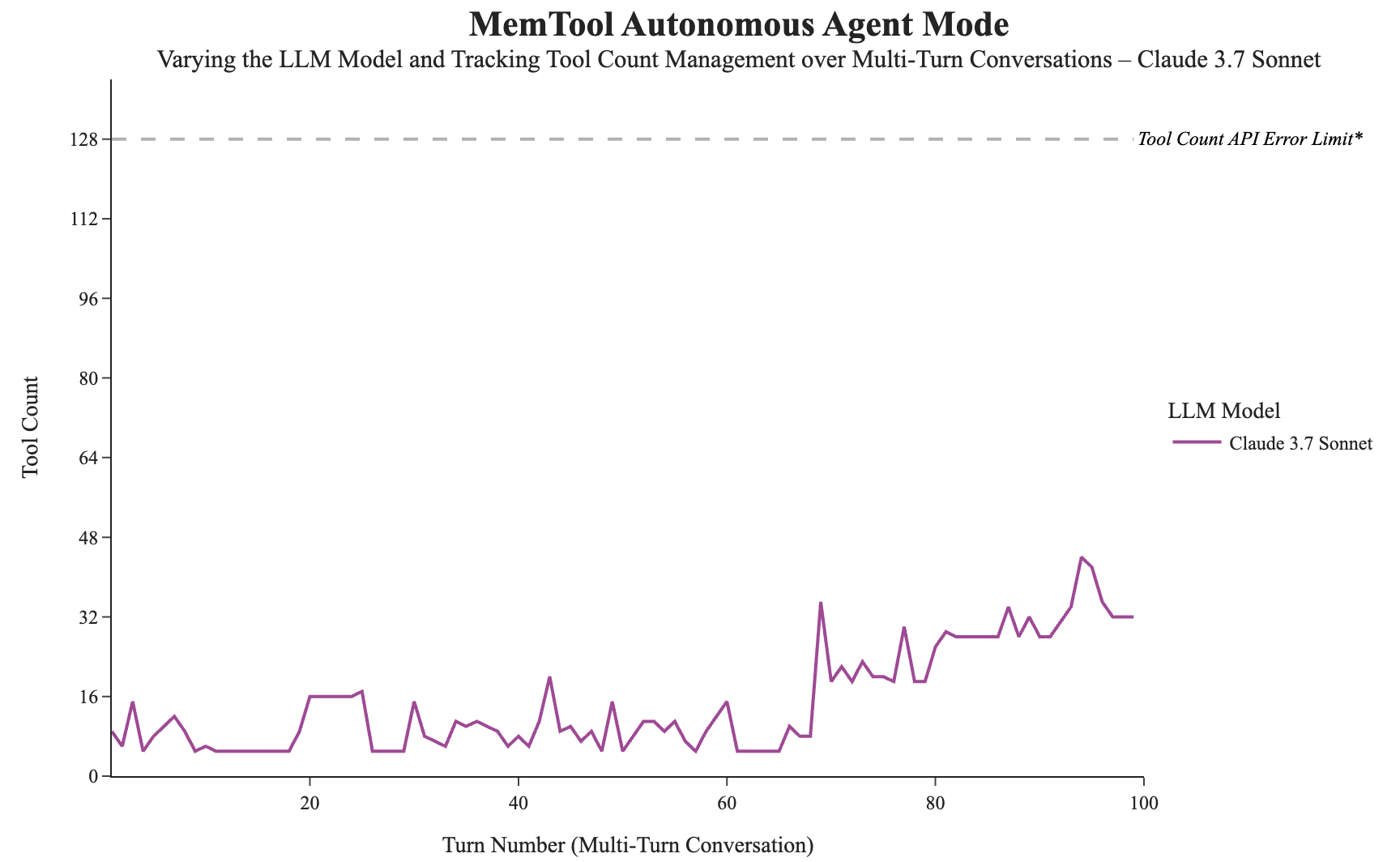}
  \caption{Claude 3.7 Sonnet — Tool Count over Time (Autonomous Agent Mode)}
  \label{fig:agent-claude37}
\end{figure*}

\begin{figure*}[t]
  \centering
  \includegraphics[width=0.8\textwidth]{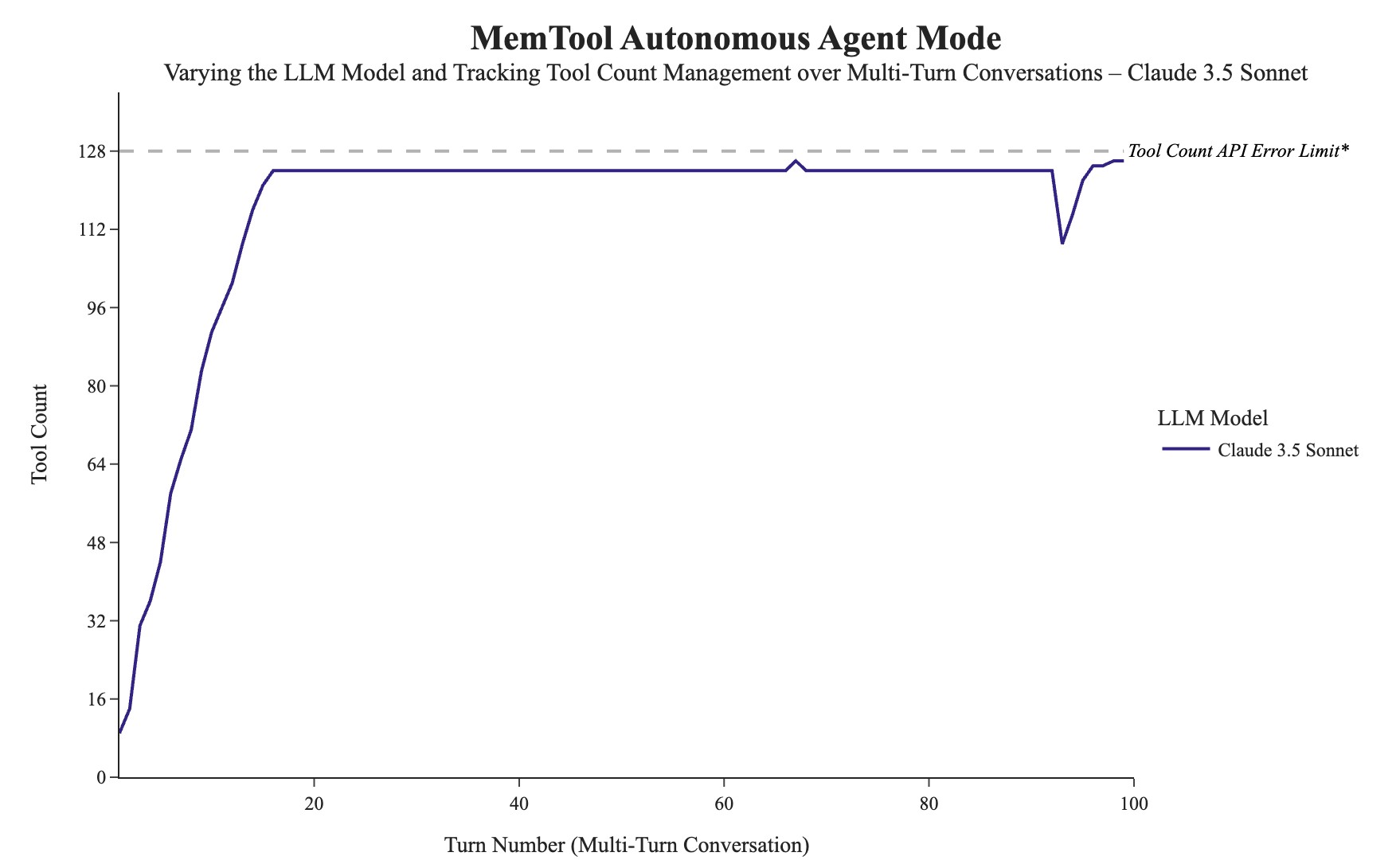}
  \caption{Claude 3.5 Sonnet — Tool Count over Time (Autonomous Agent Mode)}
  \label{fig:agent-claude35}
\end{figure*}

\begin{figure*}[t]
  \centering
  \includegraphics[width=0.8\textwidth]{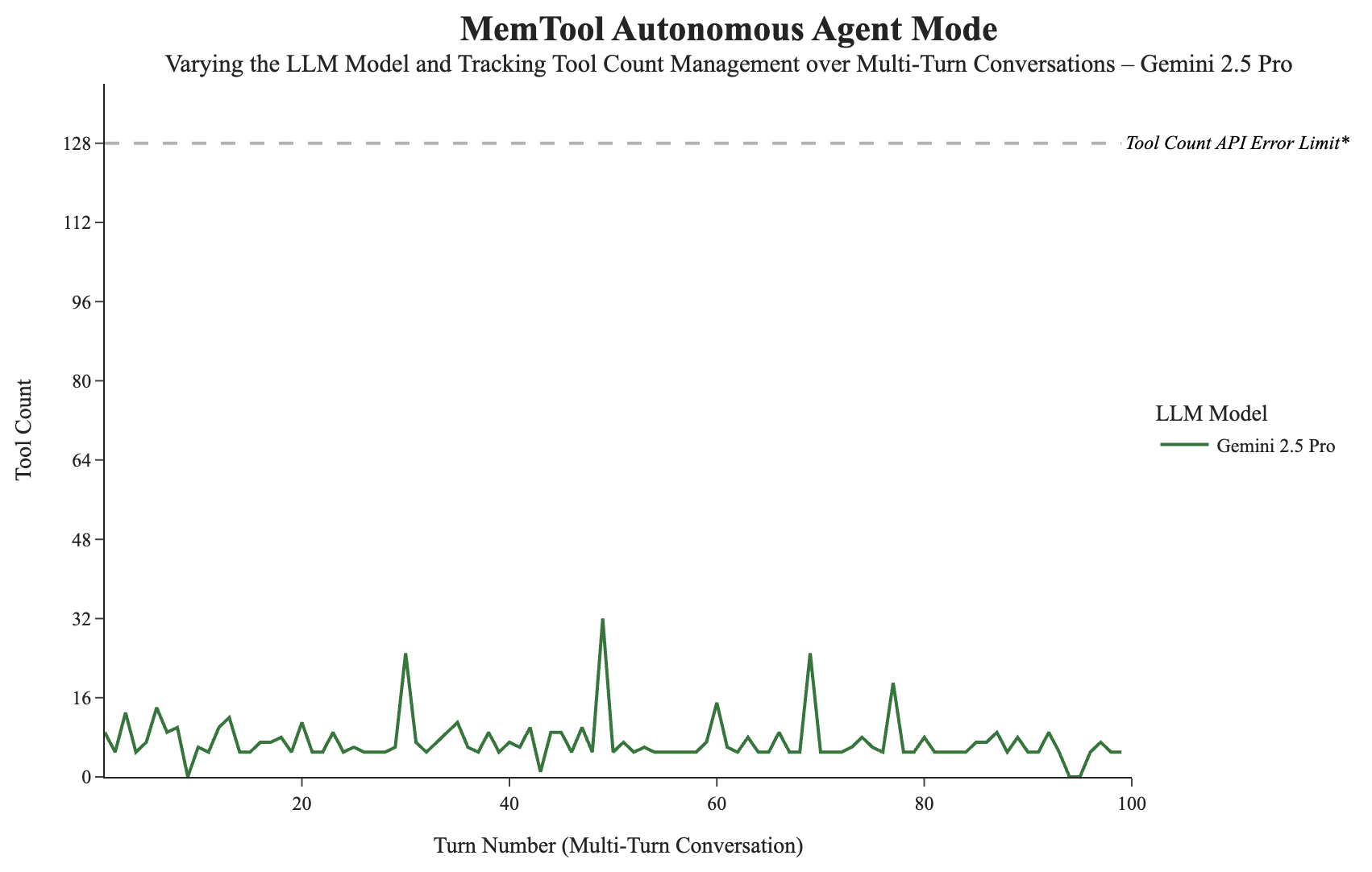}
  \caption{Gemini 2.5 Pro — Tool Count over Time (Autonomous Agent Mode)}
  \label{fig:agent-gemini25pro}
\end{figure*}

\begin{figure*}[t]
  \centering
  \includegraphics[width=0.8\textwidth]{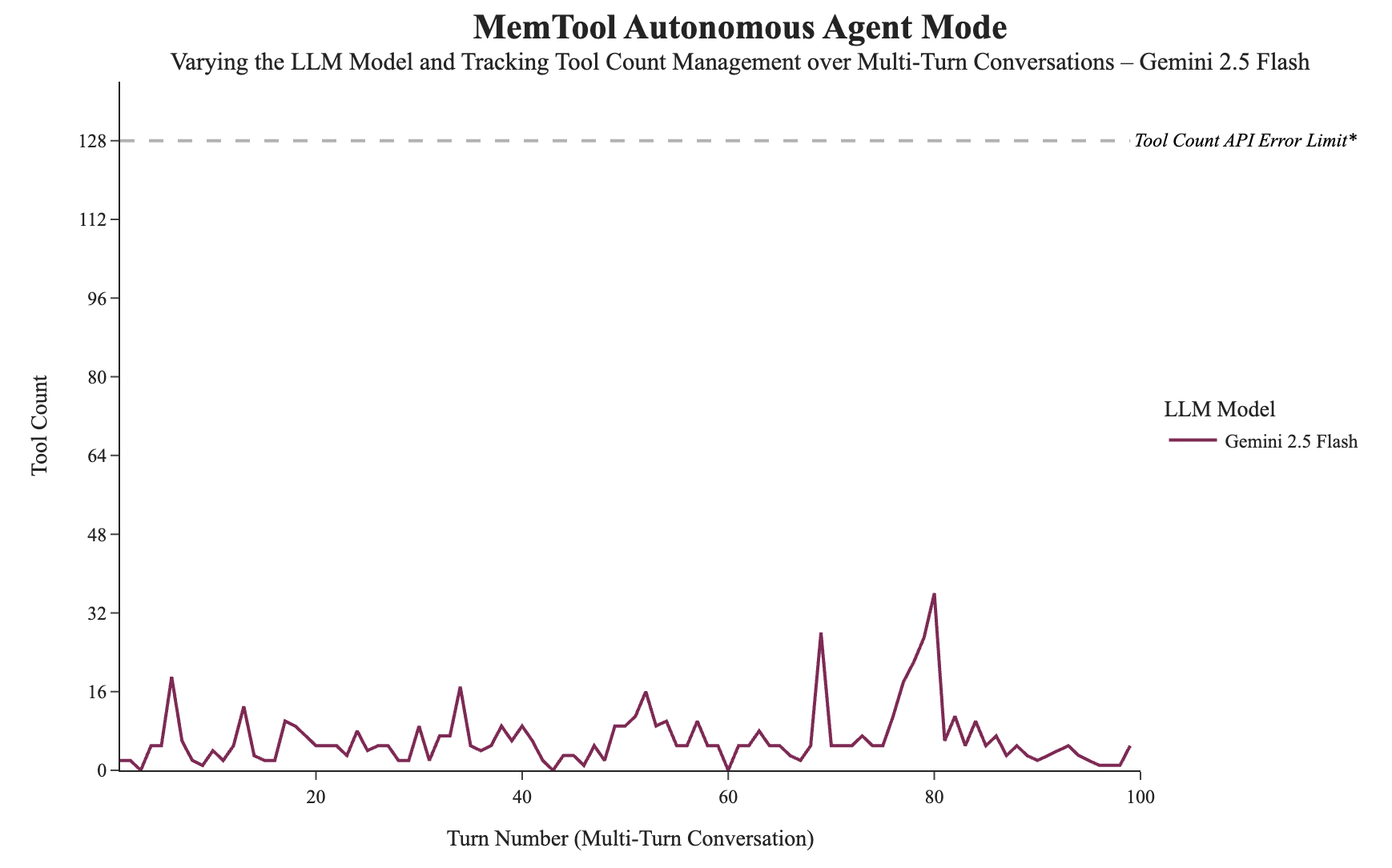}
  \caption{Gemini 2.5 Flash — Tool Count over Time (Autonomous Agent Mode)}
  \label{fig:agent-gemini25flash}
\end{figure*}

\begin{figure*}[t]
  \centering
  \includegraphics[width=0.8\textwidth]{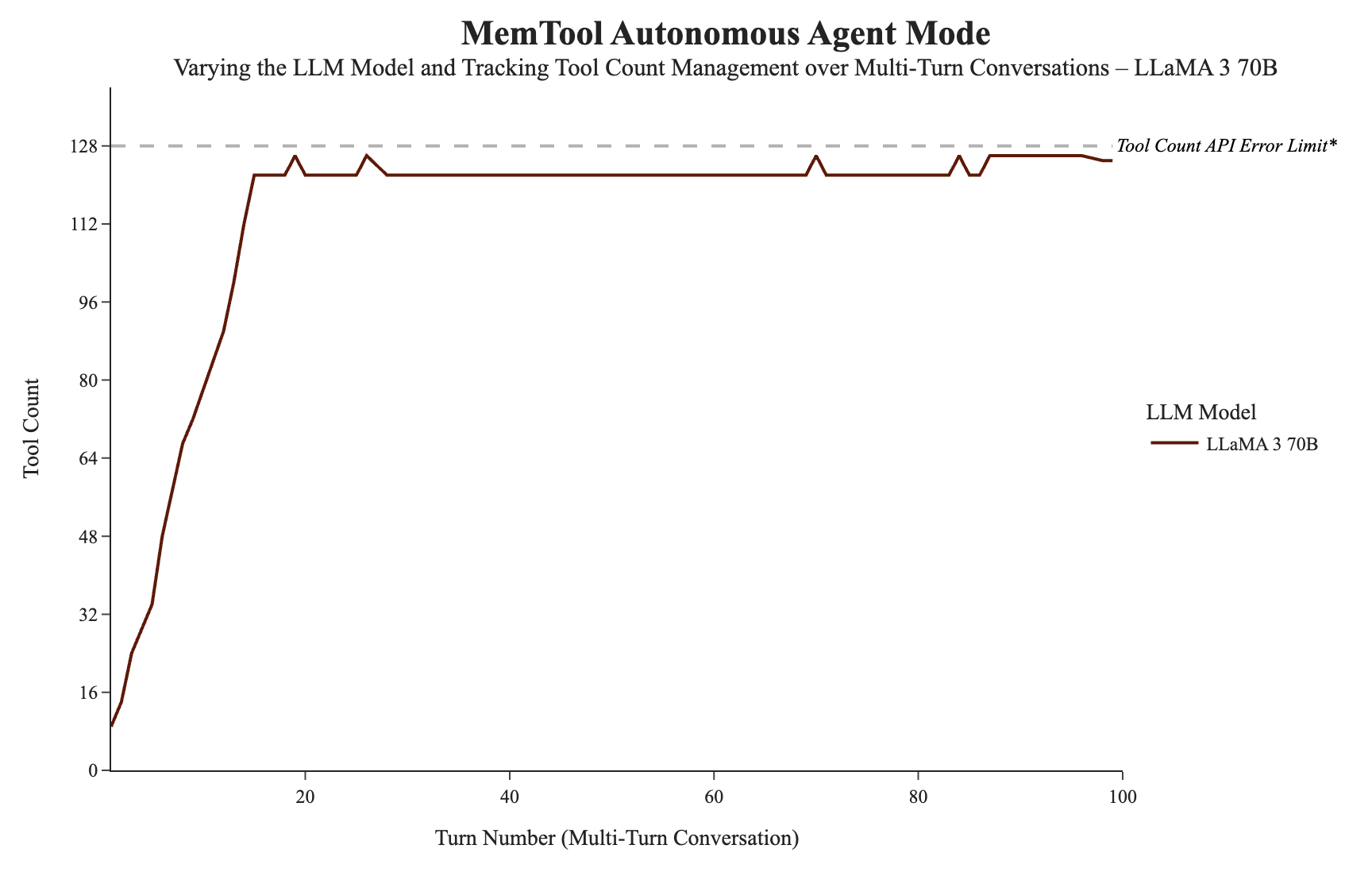}
  \caption{LLaMA 3 70B — Tool Count over Time (Autonomous Agent Mode)}
  \label{fig:agent-llama3}
\end{figure*}

\end{document}